\documentclass[journal]{IEEEtran}
\usepackage{graphicx}
\usepackage{subfigure}
\usepackage{amsmath}
\usepackage{bm}
\usepackage{algorithm}
\usepackage{algorithmic}
\usepackage{booktabs}
\usepackage[utf8]{inputenc} 
\usepackage[T1]{fontenc}    
\usepackage[hidelinks]{hyperref}       
\usepackage{url}            
\usepackage{booktabs}       
\usepackage{amsfonts}       
\usepackage{nicefrac}       
\usepackage{microtype}      
\usepackage{amsthm}
\usepackage{amssymb}
\usepackage{xspace}
\usepackage{xfrac}
\usepackage{subfigure}

\newcommand*{\eg}{e.g.\@\xspace}
\newcommand*{\ie}{i.e.\@\xspace}

\makeatletter
\newcommand*{\etc}{%
    \@ifnextchar{.}%
        {etc}%
        {etc.\@\xspace}%
}

\makeatother

\DeclareMathOperator*{\argmin}{\arg\!\min}

\newcommand{\bs}[1]{\boldsymbol{#1}} 
\renewcommand{\vec}[1]{\mathbf{#1}}
\newcommand{\mat}[1]{\mathbf{#1}}
\newcommand{\x}[0]{\vec{x}}

\newcommand{\z}[0]{\vec{z}}
\newcommand{\J}[0]{\mat{J}}

\newcommand{\dif}[1]{\mathrm{d}#1} 
\newcommand{\KL}[2]{\text{KL}({#1}||{#2})} 
\newcommand{\T}[0]{^{\top}}

\newcommand{\xtx}[1]{#1\T \mkern-1.5mu\relax #1} 

\def\R{\mathbb{R}}  
\def\E{\mathbb{E}}  
\def\N{\mathcal{N}} 

\begin{document}
%
\title{Geodesic Clustering in Deep Generative Models}
%
%
%

\author{Tao Yang,
        Georgios Arvanitidis,
        Dongmei Fu,
        Xiaogang Li,
        and S{\o}ren Hauberg
\thanks{T.~Yang, D.~Fu and X.~Li are at the University of Science and Technology Beijing, China.
 E-mail: \texttt{\{yangtao,fdm\_ustb\}@ustb.edu.cn},\quad \texttt{lixiaogang99@263.net} }
\thanks{G.~Arvanitidis and S.~Hauberg are with the Technical University of Denmark.
 E-mail: \texttt{\{gear, sohau\}@dtu.dk}.}
}

%
%

\markboth{}%
{}
%



\maketitle

\begin{abstract}
  Deep generative models are tremendously successful in learning low-dimensional
  latent representations that well-describe the data. These representations, however,
  tend to much distort relationships between points, \ie pairwise distances
  tend to not reflect semantic similarities well. This renders unsupervised
  tasks, such as clustering, difficult when working with the latent representations.
  We demonstrate that taking the geometry of the generative model into account
  is sufficient to make simple clustering algorithms work well over latent representations.
  Leaning on the recent finding that deep generative models constitute stochastically
  immersed Riemannian manifolds, we propose an efficient algorithm for computing
  geodesics (shortest paths) and computing distances in the latent space, while taking its
  distortion into account. We further propose a new architecture for modeling
  uncertainty in variational autoencoders, which is essential for understanding
  the geometry of deep generative models.
  Experiments show that the geodesic distance is very likely to
  reflect the internal structure of the data.
\end{abstract}


%
\IEEEpeerreviewmaketitle

\section{Introduction}
%
%
%
%
%
\IEEEPARstart{U}{nsupervised} learning is generally considered one of the greatest
challenges of machine learning research. In recent years, there has been a great
progress in modeling data distributions using deep generative models \cite{NIPS2014_5352,NIPS2015_5773},
and while this progress has influenced the clustering literature, the full potential
has yet to be reached.

\begin{figure}
  \centering
  \includegraphics[width=0.4\textwidth]{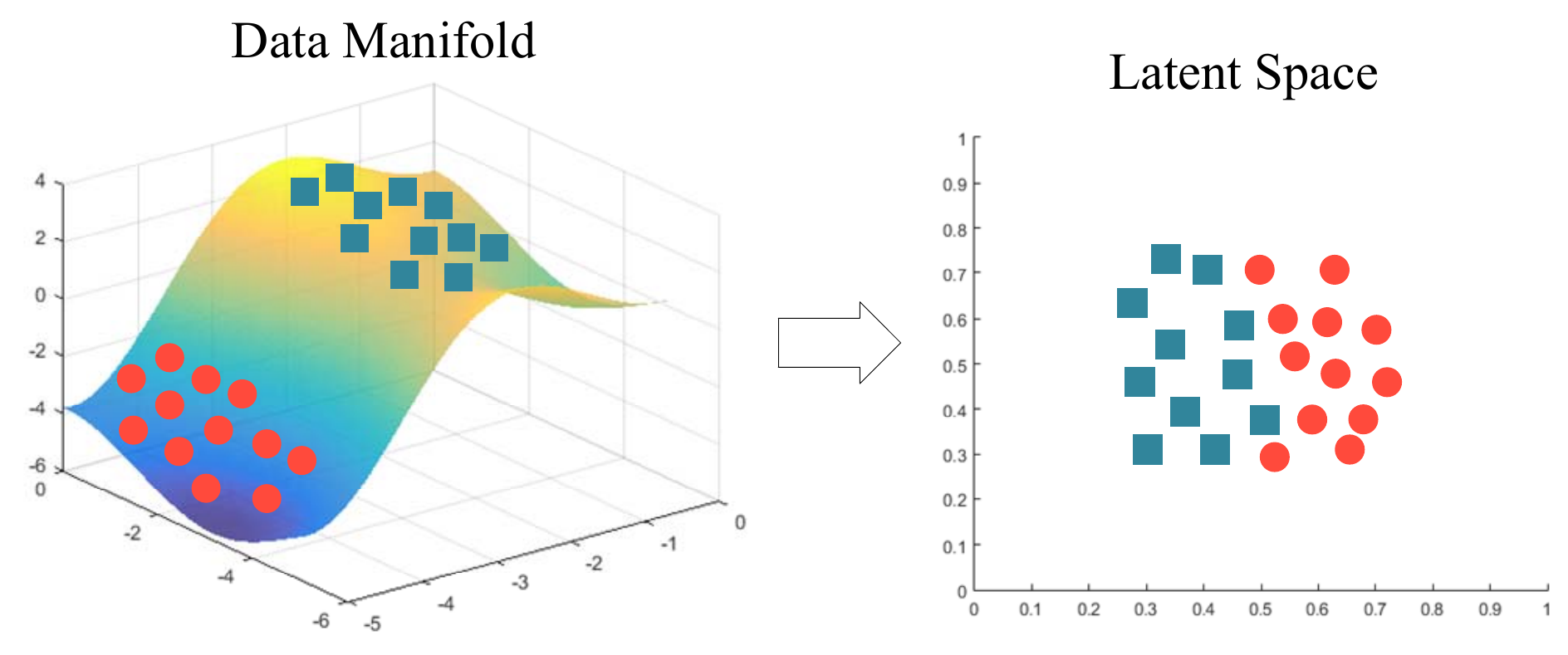}
  \caption{The latent space of a generative model is highly distorted and will
    often loose clustering structure.}\label{fig:teaser}
\end{figure}

Consider a latent variable model
\begin{align}\label{eq:lvm}
  p(\x) = \int p(\x|\z) p(\z) \dif{\z},
\end{align}
where \emph{latent variables} $\z \in R^d$ provide a low-dimensional representation of
data $\x \in R^D$ and $D\gg d$. In general, the prior $p(\z)$ will determine if clustering
of the latent variables is successful; \eg the common Gaussian prior, $p(\z) = \N(\z| \vec{0}, \mat{I_d})$,
tend to move clusters closer together, making \textit{post hoc} clustering difficult
(see Fig.~\ref{fig:teaser}).
This problem is particularly evident in deep generative models such as
\emph{variational autoencoders (VAE)} \cite{diederik14_vae, danilo14_vae} that pick
\begin{align}\label{eq:VAE}
  p(\x|\z;\theta) = \N (\x | \bm{\mu}(\z;\theta),\mat{I}_D \cdot \bm{\sigma}^2(\z;\theta)),
\end{align}
where the mean $\bm{\mu}$ and variance $\bm{\sigma}^2$ are parametrized by
deep neural networks with parameters $\theta$. The flexibility of such networks
ensures that the latent variables can be made to follow almost any prior $p(\z)$,
implying that the latent variables can be forced to show almost any structure,
including ones not present in the data. This does not influence the distribution
$p(\x)$, but it can be detrimental for clustering.

These concerns indicate that one should be very careful when computing distances
in the latent space of deep generative models. As these models (informally)
span a manifold embedded in the data space, one can consider measuring
distances \emph{along} this manifold; an idea that share intuitions with
classic approaches such as \emph{spectral clustering} \cite{spectralClus}.
Arvanitidis et al.\ \cite{georgios18} have recently shown that measuring
\emph{along} the data manifold associated with a deep generative model can be
achieved by endowing the latent space with a Riemannian metric and measure
distances accordingly. Unfortunately, the approach of Arvanitidis et al.\ require
numerical solutions to a system of ordinary differential equations, which cannot
be readily evaluated using standard frameworks for deep learning (see Sec.~\ref{sec:related}).
In this paper, we propose an efficient algorithm for evaluating these distances and
demonstrate usefulness for clustering tasks.

\section{Related Work}\label{sec:related}
Clustering, as a fundamental problem in machine learning, highly depends on the
quality of data representation. Recently deep neural networks have become
useful in learning clustering-friendly representations.
We see four categories of work based on network structure:
\emph{autoencoders (AE)}, \emph{deep neural networks (DNN)},
\emph{generative adversarial networks (GAN)} and \emph{variational autoencoders (VAE)}.

\textbf{In AE-based methods}, \emph{Deep Clustering Networks} \cite{dcn} directly
combine the loss functions in autoenconders and \textit{k}-means, while
\emph{Deep Embedding Network} \cite{den} also revised the loss function by adding
locality-preserving and group sparsity constraints to guide the network for
clustering.
\emph{Deep Multi-Manifold Clustering} \cite{dmc} introduced manifold locality
preserving loss and proximity to cluster centroids, while \emph{Deep Embedded
Regularized Clustering} \cite{depict} established a non-trivial structure of
convolutional and softmax autoencoder and proposed an entropy loss function with
clustering regularization. \emph{Deep Continuous Clustering} \cite{dcc} inherited
the continuity property in the method of \emph{Robust Continuous
Clustering}\cite{rcc}, a formulation having a clear continuous objective and no prior knowledge of clusters number, to integrate the
parameters learning in network and clustering altogether. The AE-based methods
are easy to implement but introduce hyper-parameters to the loss and are very limited in network depth.

\textbf{In DNN-based methods}, the networks can be very flexible such as
\emph{Convolutional Neural Networks} \cite{cnn_classic} or \emph{Deep Belief Networks}
\cite{dbn_classic} and often involve pre-training and fine-tuning stages.
\emph{Deep Nonparametric Clustering} \cite{dnc_arx} and \emph{Deep Embedded
Clustering} \cite{dec_xie} are such representative works. Since the network
initialization is sensitive to the result, \emph{Clustering Convolutional Neural
Networks} \cite{ccnn} is proposed with initial cluster centroids. To get rid of
pre-training, \emph{Joint Unsupervised Learning} \cite{jule} and
\emph{Deep Adaptive Image Clustering} \cite{dac_cvpr} are proposed for hierarchical
cluster and binary relationship of images specifically.

\textbf{In VAE-based methods}, because the VAE is a generative model,
\emph{Variational Deep Embedding} \cite{vade} and \emph{Gaussian Mixture VAE}
\cite{gmVAE} designed special prior distributions over the latent representation
and inferred the data classes correspond to the modes of different priors.

\textbf{In GAN-based methods}, as another generative model, \emph{Deep Adversarial
Clustering} \cite{gan_dac} was inspired by the ideas behind the \emph{Variational
Deep Embedding} \cite{vade}, but with GAN structure. \emph{Information Maximizing
Generative Adversarial Network} \cite{infogan} can disentangle the latent
representations both discrete and continuous, and it modeled a clustering function
with categorical values for those latent codes.
AE-based and DNN-based methods
are designed specifically for clustering but do not consider the underlying structure of data resulting in having no ability to generate data.
VAE-based and GAN-based methods can generate samples and infer the structure of data while because of changing the latent space for clustering, it may conversely affect the true intrinsic structure of data.

\textbf{Our work}, is based on a recent observation that deep generative models
immerse random Riemannian manifolds \cite{georgios18}. This implies a change in
the way distances are measured in the latent space, which reveals a clustering
structure. Unfortunately, practical algorithms for actually computing such distances
are missing, and it is the main focus of the present paper. With such an algorithm
in hand, clustering can be performed with high accuracy in the latent space of
an off-the-shelf VAE.

\textbf{The paper is organized as follows:}
Section~\ref{sec:VAE} introduce the usual VAE network, along with its interpretation
as a stochastic Riemannian manifold. In Sec.~\ref{sec:geodesics} we derive an efficient
algorithm for computing geodesics (shortest paths) over this manifold,
and in Sec.~\ref{sec:results} we demonstrate its usefulness for clustering tasks.
The paper is concluded in Sec.~\ref{sec:conclusion}.

\section{Background on Variational Autoencoders}\label{sec:VAE}
Deep generative modeling is an area of machine learning which deals with models
of distribution $p(\x)$ in a potentially high-dimensional space $\mathcal{X} \subseteq \R^D$.
Deep generative models can capture data dependencies by learning  low-dimensional
latent variables $\z$ to form a latent space $\mathcal{Z}\subseteq \R^d$.
In recent years, the \emph{variational autoencoder (VAE)} has emerged as one of
the most popular deep generative model because it can be built on top of deep
neural networks and be trained with stochastic gradient descent.
The VAE aims to maximize the probability of data samples $\x$ generated as
\begin{equation}\label{eq1}
  p(\x) = \int p(\x|\z;\theta)p(\z) \dif{\z}.
\end{equation}
Here, the latent variables $\z$ are sampled according to a probability density
function defined over $\mathcal{Z}$ and the distribution $p(\x|\z;\theta)$
denotes the likelihood parametrized by $\theta$. In VAEs $p(\x|\z;\theta)$ is often Gaussian
\begin{equation}\label{eq2}
  p(\x|\z;\theta) = \N (\x|\bm{\mu}(\z;\theta),\mat{I}_D \cdot \bm{\sigma}^2(\z;\theta) ),
\end{equation}
where $\bm{\mu}(\z;\theta):\R^d\rightarrow \R^D$ is the mean function and
$\bm{\sigma}^2(\z;\theta):\R^d \rightarrow \R^D_{+}$ is the covariance function.

\subsection{Inference and Generator}
The VAE consists of two parts: an \emph{inference network} and a \emph{generator network},
that serve almost the same roles as \emph{encoders} and \emph{decoders} in classic
autoencoders.

\subsubsection{The inference network}
is trained to map the training data samples $\x$ to the latent space $\mathcal{Z}$ meanwhile forcing the latent variables $\z$ to comply with the distribution $p(\z)$. However, both the posterior distribution $p(\z|\x)$ and $p(\x)$ are unknown. Therefore, VAE gives the solution that the posterior distribution is a variational distribution $q(\z|\x;\lambda)$, computed by a network with parameters $\lambda$. In order to make $q(\z|\x;\lambda)$ accord with the distribution $p(\z)$, the Kullback-Leibler (KL) divergence \cite{Bishop:2006:PRM:1162264} is used, that is:
\begin{equation}\label{eq3}
  \min_{\lambda}\KL{q(\z|\x;\lambda)}{p(\z)}
\end{equation}

\subsubsection{The generator network}
is trained to map the latent variables $\z$ to generate data samples $\hat{\x}$ that are much like the true samples $\x$ from the data space $\mathcal{X}$. According to the purpose of this network, we know that it needs to maximize the marginal distribution $p(\x|\z;\theta)$ over the whole latent space and actually it is often processed with logarithm and computed by a multi-layer network with parameters $\theta$:
\begin{equation}\label{eq4}
  \max_{\theta} \E_{\z\sim q(\z|\x;\lambda)} [\log p(\x|\z;\theta)]
\end{equation}
From these parts a VAE is jointly trained as
\begin{equation}\label{eq5}
  \theta^*,\lambda^*=\arg\max_{\lambda,\theta} \E[\log p(\x|\z;\theta)] - \KL{q(\z|\x;\lambda)}{p(\z)} .
\end{equation}

\begin{figure}
  \centering
  \includegraphics[width=0.4\textwidth]{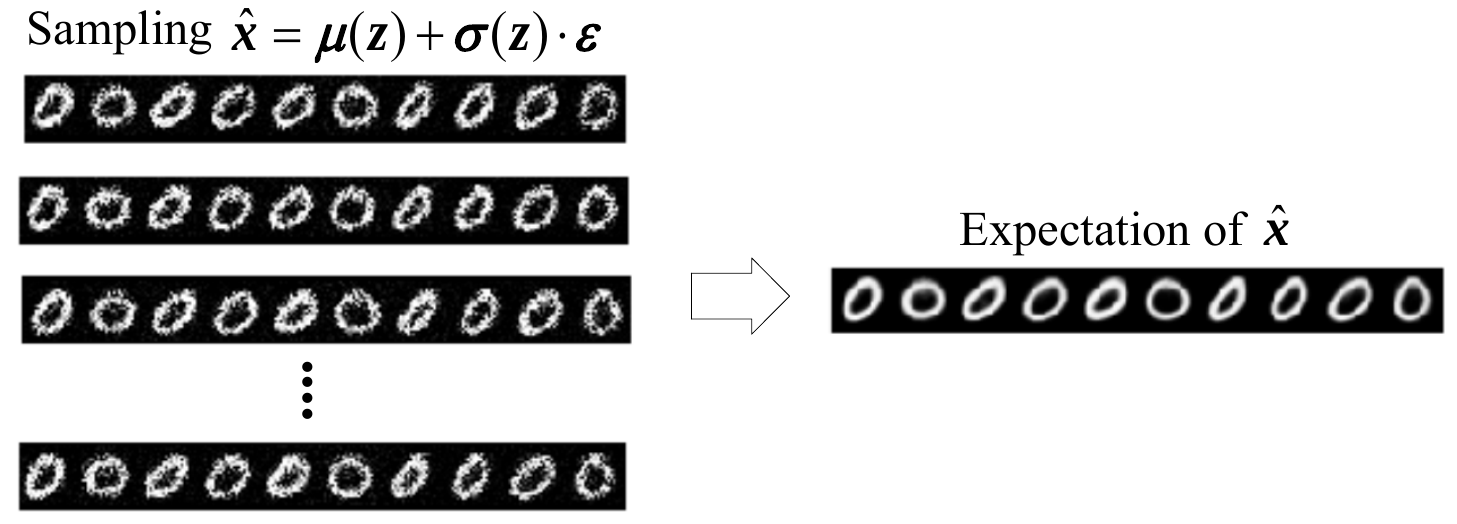}
  \caption{Expectation of samplings from distribution $p(\x|\z;\theta)$}\label{fig2}
\end{figure}

\subsection{The Random Riemannian Interpretation}
The inference network should force the latent variables to approximately
follow the pre-specified unit Gaussian prior $p(\z)$, which implies that the latent
space gives a highly distorted view of the original data. Fortunately, this
distortion is fairly easy to characterize \cite{georgios18}. First, observe that the
generative model of the VAE can be written as (using the so-called \emph{re-parametrization trick}; see also Fig.~\ref{fig2})
\begin{equation}\label{eq:reparam_trick}
  \hat{\x} = f(\z) = \bm{\mu}(\z) + \bm{\sigma}(\z) \odot \bm{\varepsilon},
  \qquad \bm{\varepsilon} \sim \N(\vec{0}, \mat{I}).
\end{equation}
Now let $\z$ be a latent
variable and let $\bs{\delta}$ be infinitesimal. Then we can measure the distance
between $\z$ and $\z + \bs{\delta}$ in the input space using Taylor's theorem
\begin{align}
  \| f(\z) - f(\z + \bs{\delta}) \|^2
    &= \| \J_{\z}\z - \J_{\z}(\z + \bs{\delta}) \|^2 \\
    &= \bs{\delta}\T \xtx{\J_{\z}} \bs{\delta},
\end{align}
where $\J_{\z}$ denote the Jacobian of $f$ at $\z$. This implies that $\xtx{\J_{\z}}$ define
a local inner product under which we can define curve lengths through integration
\begin{align}
  \text{Length}(\vec{c})
    &= \int_a^b \| \partial_t f(\vec{c}_t) \| \dif{t}
     = \int_a^b \sqrt{ \dot{\vec{c}}_t\T \xtx{\J_{\z}} \dot{\vec{c}}_t} \dif{t}.
  \label{eq:length}
\end{align}
Here $\vec{c}: [a, b] \rightarrow \R^d$ is a curve in the latent space and
$\dot{\vec{c}} = \partial_t \vec{c}$ is its velocity. Distances can then be
defined as the length of the shortest curve (geodesic) connecting two points,
\begin{align}
  \text{dist}(\z_0, \z_1)
    &= \text{Length}\left( \vec{c}^{(0,1)} \right) \\
  \vec{c}^{(0,1)}
    &= \argmin_{\substack{\vec{c},\ \vec{c}_0 = \z_0,\\ \vec{c}_1 = \z_1}} \text{Length}(\vec{c})
\end{align}
This is the traditional Riemannian analysis associated with embedded surfaces \cite{gallot1990riemannian}.
From this, it is well-known that length-minimizing curves are minimizers of \emph{energy}
\begin{align}
  \mathcal{E}(\vec{c})
    = \int_a^b \dot{\vec{c}}_t\T \xtx{\J_{\z}} \dot{\vec{c}}_t \dif{t},
\end{align}
which is easier to optimize than Eq.~\ref{eq:length}.

For generative models, the analysis is complicated by the fact that $f$ is a stochastic
mapping, implying that the Jacobian $\J_{\z}$ is stochastic, geodesics are stochastic,
distances are stochastic, \etc. Arvanitidis et al.\ \cite{georgios18} propose
to replace the stochastic metric $\xtx{\J_{\z}}$ with its expectation $\E[\xtx{\J_{\z}}]$
which is equivalent to minimizing the expected energy \cite{hauberg:only:2018}.
While this is shown to work well, the practical algorithm proposed by Arvanitidis et al.\
amount to solving a nonlinear differential equation numerically, which require
us to evaluate both the Jacobian $\J_{\z}$ and its derivatives. Unfortunately,
modern deep learning frameworks such as Tensorflow rely on
\emph{reverse mode} automatic differentiation \cite{rall1981automatic}, which does not support
Jacobians. This renders the algorithm of Arvanitidis et al.\ impractical.
A key contribution of this paper is a practical algorithm for computing
geodesics that fits within modern deep learning frameworks.

\begin{figure}
  \centering
  \includegraphics[width=0.4\textwidth]{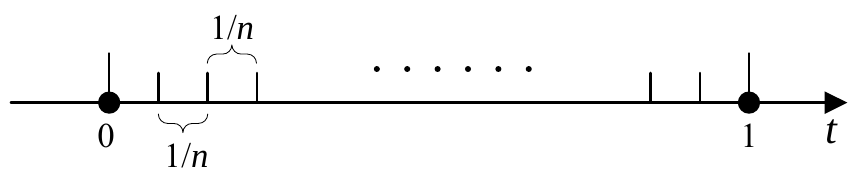}
  \caption{Discretization of parameter $t$}\label{fig4}
\end{figure}

\section{Proposed algorithm to compute geodesics}\label{sec:geodesics}
  To develop an efficient algorithm for computing geodesics, we first note that
  the expected curve energy can be written as
  \begin{align}
    \bar{\mathcal{E}}
      &= \E[\mathcal{E}(\vec{c})]
       = \int_a^b \E\left[ \| \partial_t f(\vec{c}_t) \|^2 \right] \dif{t}.
  \end{align}
  If we discretize the curve $\vec{c}$ at $n$ points (Fig.~\ref{fig4}), then this
  integral can be approximated as
  \begin{align}
    \bar{\mathcal{E}}
      &\approx \sum_{i=0}^{n-1} \E\left[ \left\| \frac{f(\vec{c}_i) - f(\vec{c}_{i+1})}{t_i-t_{i+1}} \right\|^2 \right] .
  \end{align}
  Since $f(\vec{c}) \sim \N(\bs{\mu}(\vec{c}), \mat{I}_D \cdot \bs{\sigma}^2(\vec{c}))$,
  the expectation computing can be evaluated in closed-form as
  \begin{align}
  \begin{split}
    \E\left[ \| f(\vec{c}_i) - f(\vec{c}_{i+1}) \|^2 \right]
      &= \big( \bm{\mu}(\vec{c}_i)      - \bm{\mu}(\vec{c}_{i+1}) \big)^2  \\
      &+ \big( \bm{\sigma}^2(\vec{c}_i) + \bm{\sigma}^2(\vec{c}_{i+1}) \big),
  \end{split}
  \end{align}
  and the approximated expected energy can be written
  \begin{align}
    \bar{\mathcal{E}}
      &\approx \sum_{i=0}^{n-1} \left\{\!
          \big( \bm{\mu}(\vec{c}_i)      \! - \! \bm{\mu}(\vec{c}_{i+1}) \big)^2
        + \big( \bm{\sigma}^2(\vec{c}_i) \! + \! \bm{\sigma}^2(\vec{c}_{i+1}) \big)
        \!\right\} \!.
    \label{eq:expE}
  \end{align}
  This energy is easily interpretable: the first term of
  the sum corresponds to the curve energy along the expected data manifold,
  while the second term penalizes curves for traversing highly uncertain regions
  on the manifold. This implies that geodesics will be attracted to regions of high
  data density in the latent space.

  Unlike the ordinary differential equations of Arvanitidis et al.\ \cite{georgios18},
  Eq.~\ref{eq:expE} can readily be optimized using automatic differentiation as
  implemented in Tensorflow. We can, thus, compute geodesics by picking
  a parametrization of the latent curve $\vec{c}$ and optimize Eq.~\ref{eq:expE}
  with respect to curve parameters.

  \subsection{Curve Parametrization}
    There are many common choices for parametrizing curves, \eg \emph{splines} \cite{georgios18},
    \emph{Gaussian processes} \cite{hennig:aistats:2014} or \emph{point collections}
    \cite{laine2018feature}. In the interest of speed, we propose to use the
    restricted class of \emph{quadratic functions}, \ie
    \begin{align}\label{eq12}
      \vec{c}_t
        &= \left[ \begin{array}{c}
             a_1 t^2 + b_1 t + c_1 \\
             a_2 t^2 + b_2 t + c_2 \\
             \vdots                \\
             a_d t^2 + b_d t + c_d \\
           \end{array}
         \right],
      \qquad
      \vec{c}: [0, 1] \rightarrow \R^d.
    \end{align}
    A curve, thus, has $3d$ free parameters $a_{:}$, $b_{:}$, and $c_{:}$.
    In practice, we are concerned with geodesic curves that connect two
    pre-specified points $\z_0$ and $\z_1$, so the quadratic function
    should be constrained to satisfy $\vec{c}_0 = \z_0$ and $\vec{c}_1 = \z_1$,
    which is easily achieved for quadratics. Under this constraint, there are
    only $d$ free parameters to estimate when optimizing $\bar{\mathcal{E}}$ \eqref{eq:expE}.
    Here we perform the optimization using standard gradient descent.

  \subsection{Specifying Uncertainty}\label{sec:GMM}
    When training the VAE model, the reconstruction term of Eq.~\ref{eq5} ensure
    that we can expect high-quality reconstructions of the training data.
    Interpolations between latent training data usually give high-quality
    reconstructions in densely sampled regions of the latent space, but low-quality
    reconstructions in regions with low sample density. Ideally, the generator
    variance $\bm{\sigma}^2(\vec{z})$ should reflect this.

    From the point of view of computing geodesics, the generator variance
    $\bm{\sigma}^2(\vec{z})$ is important as it appears directly in the
    expected curve energy \eqref{eq:expE}. If $\bm{\sigma}^2(\vec{z})$ is
    small near the latent data and large away from the data, then geodesics
    will follow the trend of the data \cite{hauberg:only:2018}, which is a useful
    property in a clustering context.

    In practice, the neural network used to model $\bm{\sigma}^2(\vec{z})$ is
    only trained where there is data, and its behavior in-between is governed
    by the activation functions of the network; \eg common activations such
    softplus or $\tanh$ implying that variances are smoothly interpolated in-between
    the training data. This is a most unfortunate property for a variance function;
    \eg if the optimal variance is low at all latent training points, then
    the predicted variance will be low at all points in the latent space.
    To ensure that variance increases away from latent data, Arvanitidis et al.\
    \cite{georgios18} proposed to model the inverse variance (precision) with
    an RBF network \cite{que:aistats:2016} with isotropic
    kernels. This is reported to provide meaningful variance estimates.

\begin{figure*}
  \subfigure[]{
    \includegraphics[width=2.0in]{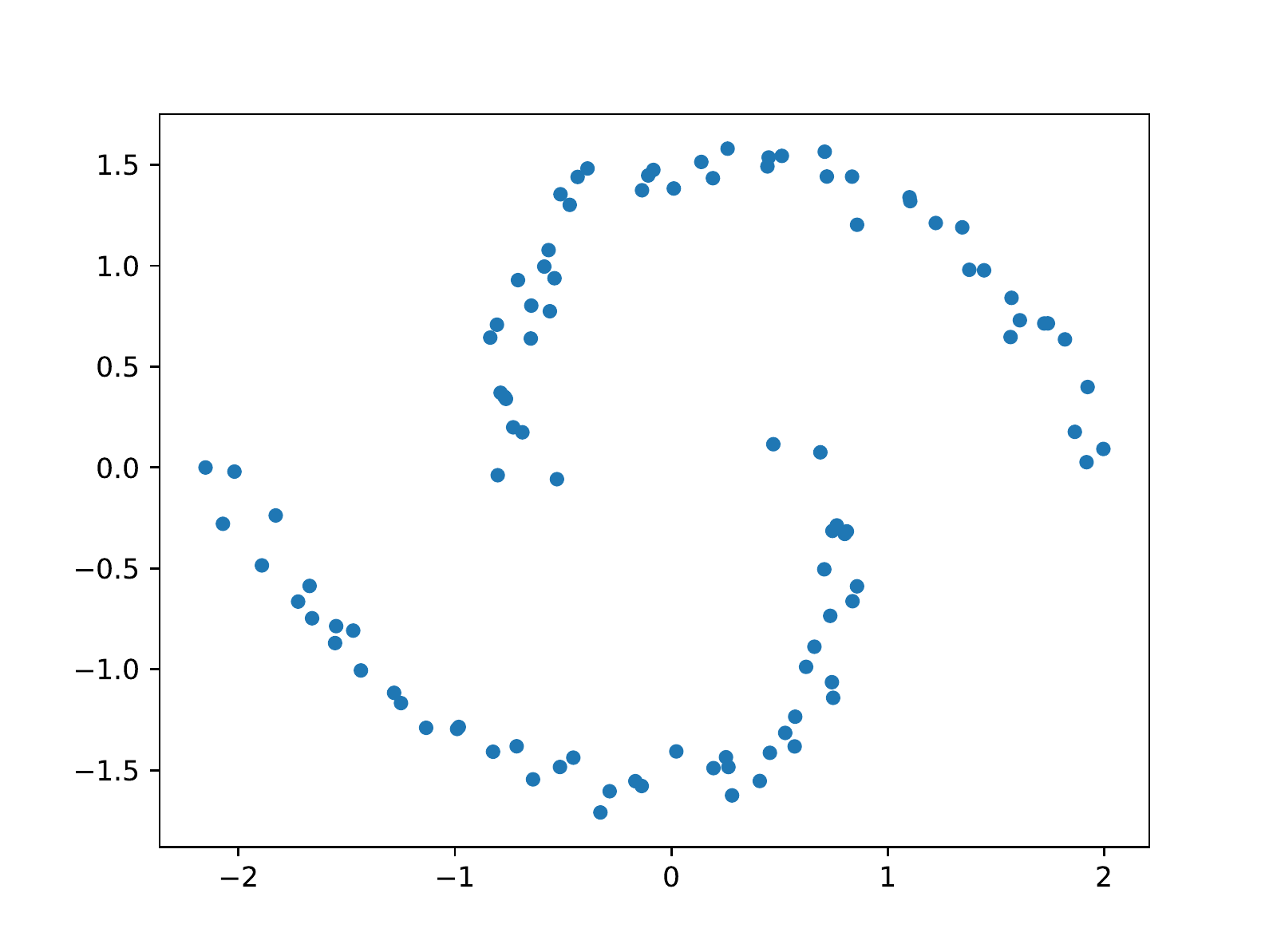}
  }
  \subfigure[]{
    \includegraphics[width=2.0in]{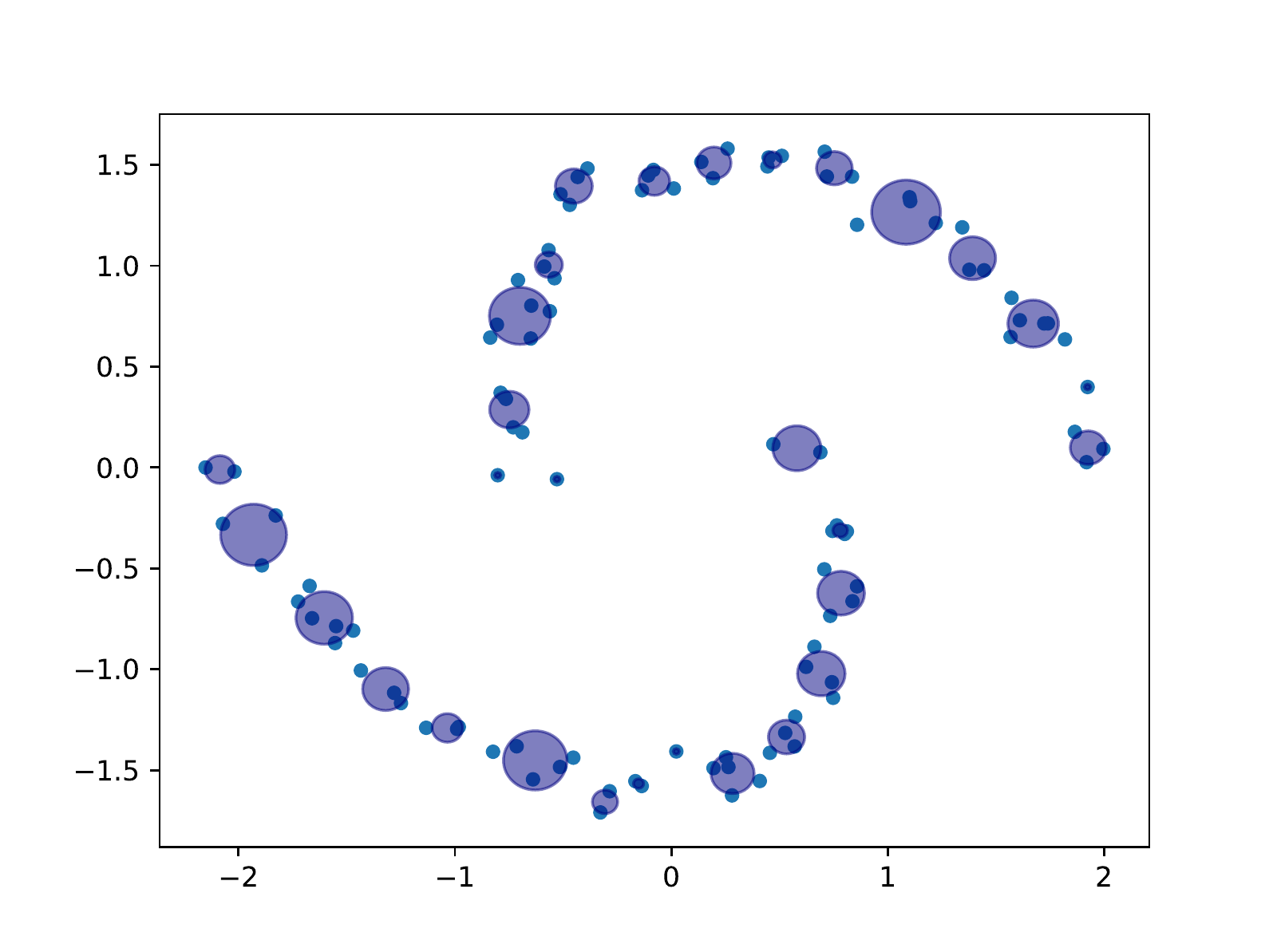}
  }
  \subfigure[]{
    \includegraphics[width=2.4in,height=1.45in]{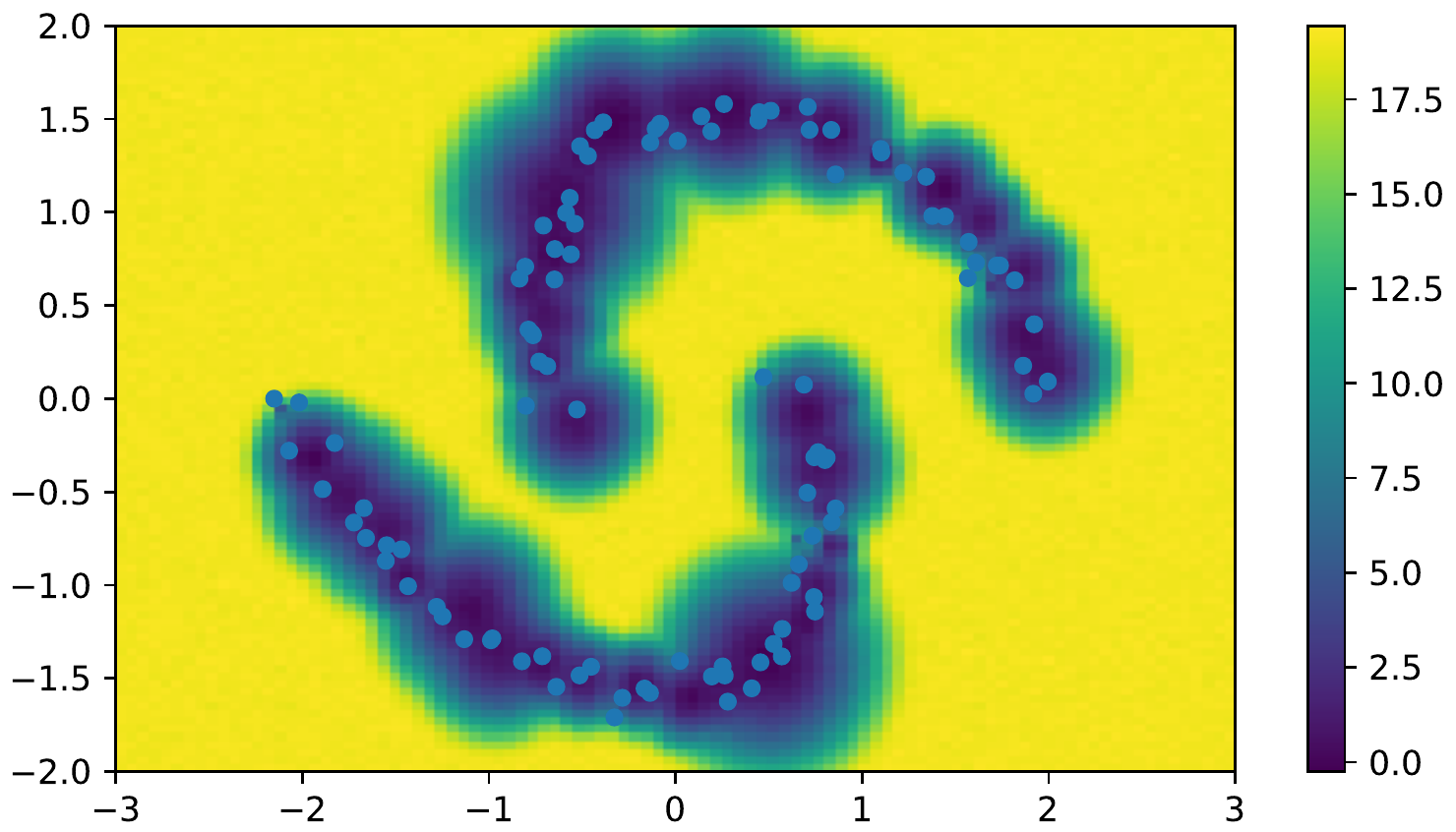}
  }
  \caption{(a): The latent samples of two-moon data. (b) The GMM result in latent space. (c) The logarithmic result of the variance.}
  \label{latent_twomoon} 
\end{figure*}

    We found the isotropic assumptions to be too limited, and instead applied
    anisotropic kernels. Specifically, we propose to use a rescaled
    Gaussian Mixture Model (GMM) to represent the inverse variance function
    \begin{align}\label{eq13}
      \frac{1}{\sigma^2(\z)} &= g(\z) = \sum_{i=1}^K w_i \N(\z ~|~ c_i, \Sigma_i) \mat{W}_g,
    \end{align}
    where $c_i$ and $\Sigma_i$ are the component-wise mean and covariance, and
    $w_i$ and $\mat{W}_g \in \R^D$ are positive weights.
    For simplicity, we set each component has its own single variance.
    For all the latent variables $\z$, we use the usual EM algorithm \cite{Bishop:2006:PRM:1162264}
    to obtain the weights $\bm{w}$ and the mean and covariance in each component.
    $\mat{W}_g$ is trained by solving Eq.~\ref{eq5}.
    Figure~\ref{latent_twomoon} gives an example, showing the inverse output of
    $g(\z)$, and we see that for regions without data, $1/g(\z)$ gives low values
    such that the variance is large as one would expect.

\begin{figure*}
  \centering
  \includegraphics[width=0.7\textwidth]{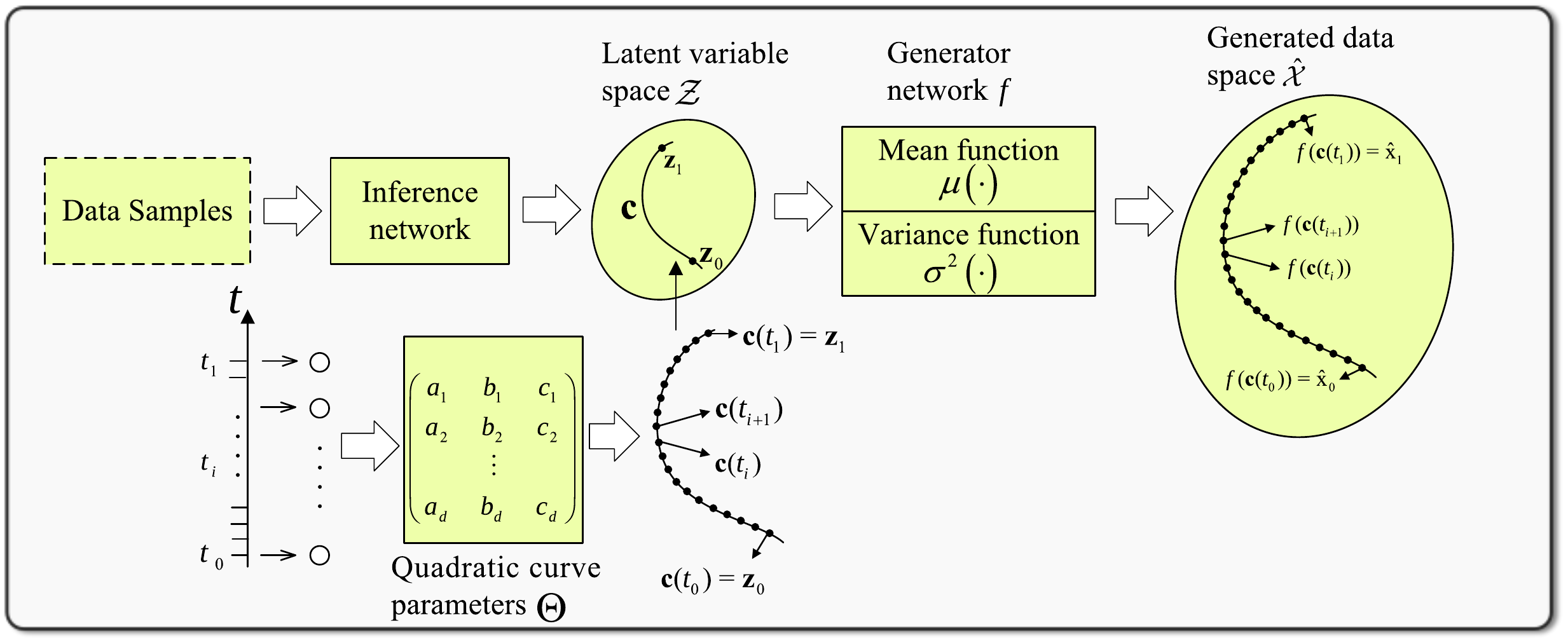}
  \caption{Construction of unified VAE and geodesic computation network}\label{vae_geo1}
\end{figure*}

\subsection{Curve Initialization}
  Once the VAE is fully trained, we can compute geodesics in the latent space.
  As previously mentioned, we use gradient descent to minimize $\bar{\mathcal{E}}$ \eqref{eq:expE}.
  To improve convergence speed, we here propose a heuristic for initialization
  that we have found to work well.

  Since geodesics generally follow the trend of the data \cite{georgios18}
  we seek an initial curve with this property. As it can be expensive to
  evaluate the generator network $f$, we propose to first seek a curve that
  minimize the inverse GMM model $g(\vec{c}) = \int g(\vec{c}_t) \dif{t}$.
  We do this with a simple stochastic optimization akin to a particle
 filter \cite{cappe2007overview}. This is written explicitly in Algorithm~\ref{InitParam}.
\begin{algorithm}[h]
  \caption{The pseudo-code for initializing $\Theta$}
  \label{InitParam}
  \begin{algorithmic}[1]
  \STATE Set $\bm{\mu}_0=0, \bm{\sigma}^2_0= \|\z_0 - \z_1\|$
  \FOR{each step in $[1,2,\cdots, \text{max\_step}]$}
    \STATE $\bm{\mu}_i=\bm{\mu}_{i-1}$, $\bm{\sigma}_i = \bm{\sigma}_{i-1}$
    \STATE Sample $M$ sets of parameters $\Theta_{1:M} \sim \mathcal{N}(\bm{\mu}_i,\mat{I_d} \cdot \bm{\sigma}^2_i )$
    \FOR{each index in $M$}
      \STATE $\vec{c}_{\text{index}} \leftarrow \vec{c}(t;\Theta_{\text{index}})$, $t\in[0,1]$
      \STATE $C_{\text{index}} \leftarrow  1/g(\vec{c}_{\text{index}})$
    \ENDFOR
    \STATE $\bm{\mu}_{i+1},\bm{\sigma}_{i+1} \leftarrow  \arg\min C_{\text{index}}$ for all indexes
  \ENDFOR
  \STATE Initialize parameters $\Theta$ from $\mathcal{N}(\bm{\mu}_{\text{max\_step}}, \mat{I_d} \cdot \bm{\sigma}^2_{\text{max\_step}})$
  \end{algorithmic}
\end{algorithm}

\section{Experiments and details related}\label{sec:results}

\subsection{Experimental Pipeline}
\begin{figure*}
  \centering
  \includegraphics[width=0.8\textwidth]{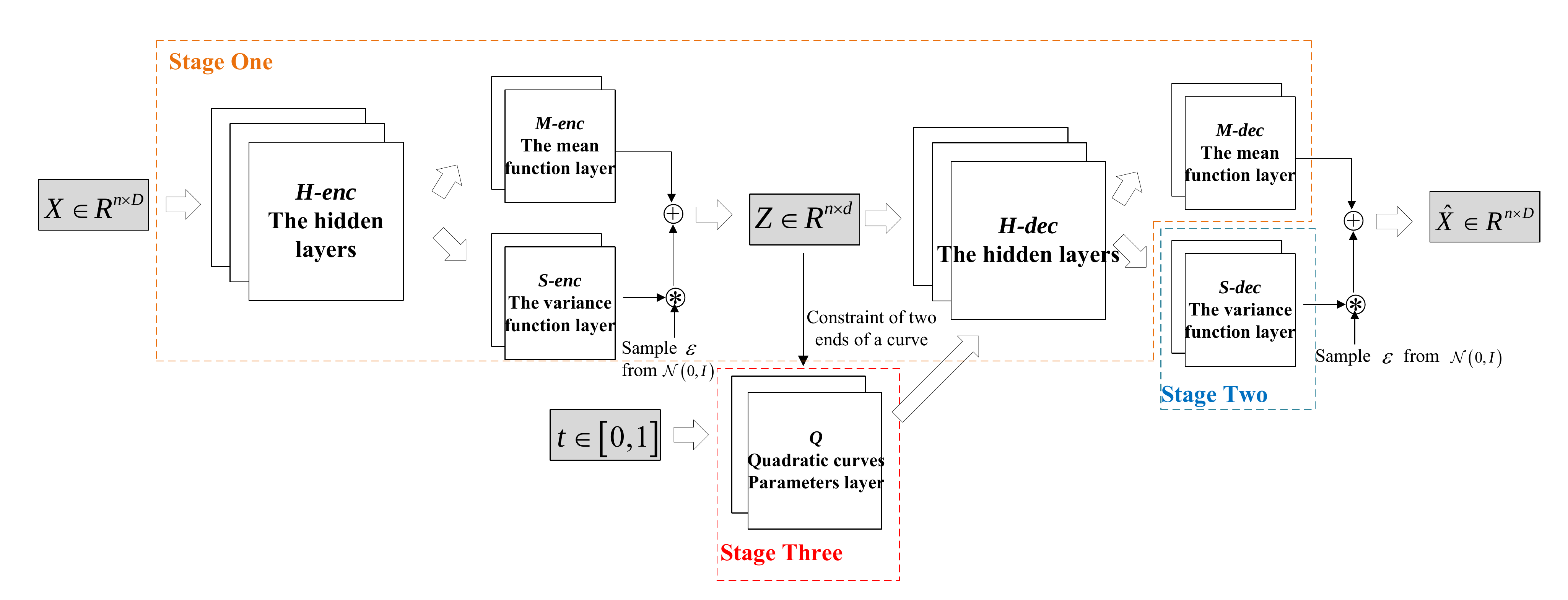}
  \caption{The three stages of our pipeline. See text for details.}\label{three_stages}
\end{figure*}

Throughout the experiments, we use the same three-stage pipeline, which is illustrated
in Fig.~\ref{three_stages}.
In the first stage we train a VAE with fixed constant output variance; this VAE
has five layers in total, \textit{H-enc}, \textit{M-enc}, \textit{S-enc}, \textit{H-dec}, \textit{M-dec}
which are optimized according to Eq.~\ref{eq5}.
In the second stage, we fit the generator variance represented by a GMM network
(Sec.~\ref{sec:GMM}) according to Eq.~\ref{eq5}.
Finally, in the third stage, we compute geodesics parametrized by $\Theta$,
and compute clusters accordingly. Here we use the $k$-medoids algorithm
\cite{kaufman1987clustering} that only rely on pairwise distances. This decision
was made to illustrate the information captured by geodesic distances.

\subsection{Visualizing Curvature}
\begin{figure}
  \centering
  \includegraphics[width=0.3\textwidth]{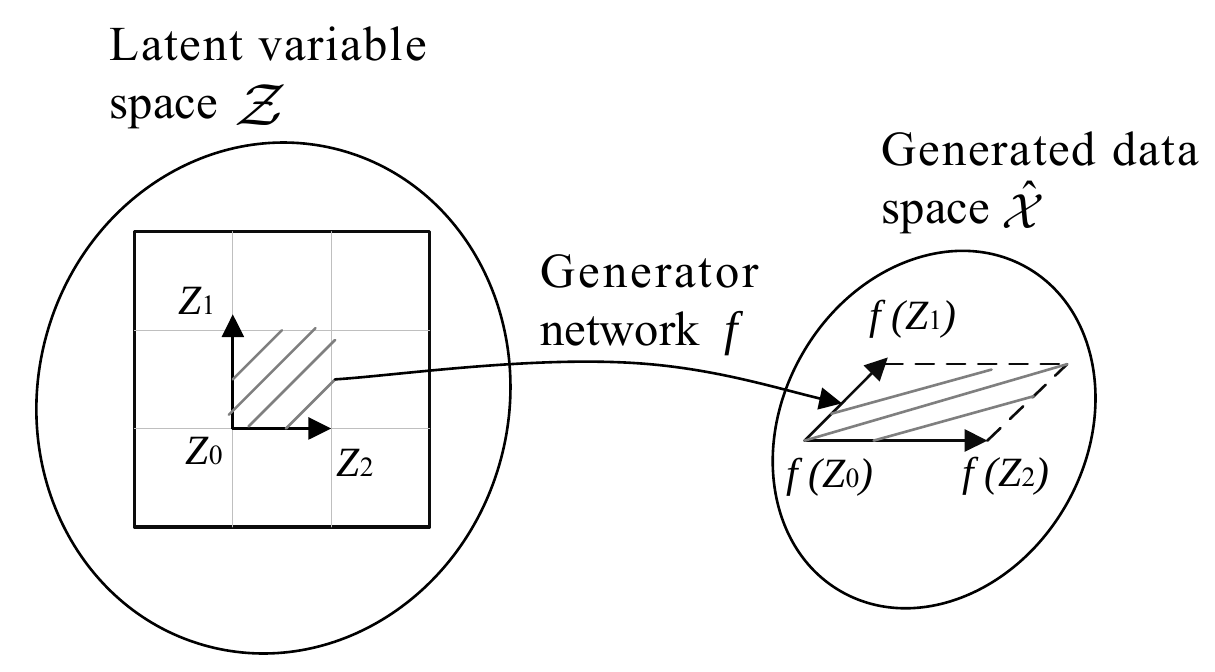}
  \caption{The volume measure describe the volume of an infinitesimal box in the latent space measured in the input space.}\label{volume_measure}
\end{figure}

\begin{figure}
  \centering
  \includegraphics[width=0.3\textwidth]{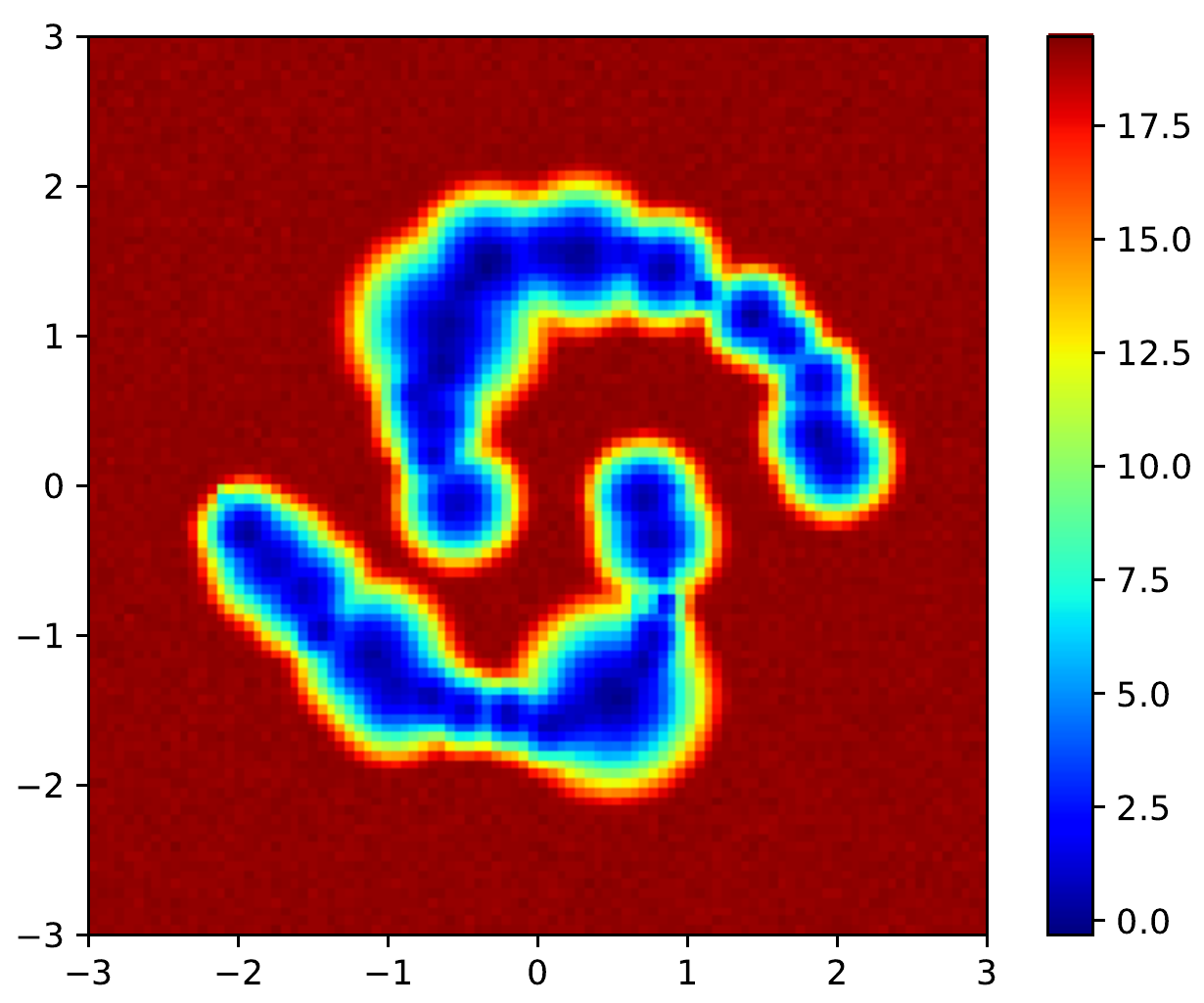}
  \caption{The logarithmic results of the volume measure}\label{twomoon_area}
\end{figure}

  A useful visualization tool for the curvature of generative models is the
  \emph{magnification factor} \cite{bishop1997magnification}, which correspond to the
  \emph{Riemannian volume measure} associated with the metric \cite{Tosi:UAI:2014}.
  For a given Jacobian $\J_{\z}$, this is defined as
  \begin{align}
    \text{vol}(\z) = \sqrt{\det \xtx{\J_{\z}}}.
  \end{align}
  In practice, the Jacobian is a stochastic object, so previous work \cite{georgios18, Tosi:UAI:2014}
  has proposed to visualize $\sqrt{\det \E[\xtx{\J_{\z}}]}$. Here we argue that
  the expectation should be taken as late in the process as possible, and instead
  visualize the expected volume measure,
  \begin{align}
    \overline{\text{vol}}(\z) = \E\left[ \sqrt{\det \xtx{\J_{\z}}} \right] .
  \end{align}
  To compute this measure, we split the latent space into small quadratic pieces,
  as in Fig.~\ref{volume_measure}.
As we can see from the figure, there are two vectors $v^{\z}_{(0,1)}=\z_1-\z_0$, $v^{\z}_{(0,2)}=\z_2-\z_0$ and the corresponding vectors in $\hat{\mathcal{X}}$, $v^{f(\z)}_{(0,1)}=f(\z_1)-f(\z_0)$ and $v^{f(\z)}_{(0,2)}=f(\z_2)-f(\z_0)$. Note $V=[v^{f(\z)}_{(0,1)},v^{f(\z)}_{(0,2)}]$, then the volume measure is:
\begin{equation}\label{eq16}
  \overline{\text{vol}}\left( v^{f(\z)}_{(0,1)},v^{f(\z)}_{(0,2)} \right) \approx \E\left[ \sqrt{\det(V^T V)} \right].
\end{equation}
Here we compute the right-hand side expectation using sampling.
As an example visualization, Fig.~\ref{twomoon_area} show the logarithm of
the volume measure associated with the model from Fig.~\ref{latent_twomoon}.
In areas of small volume measure (blue), distances will generally be small,
while they will be large in regions of large volume measure (red).

\subsection{Experimental Results}
\subsubsection{The Two-Moon Dataset}
As a first illustration, we consider the classic ``Two Moon'' data set shown in
Fig.~\ref{latent_twomoon}. For \textit{H-enc} and \textit{H-dec} layers, we use
two hidden fully-connected layers with softplus activations, and for the
\textit{S-enc} layer, we use one fully-connected layer, again, with softplus.
For \textit{M-enc} and \textit{M-dec} layers we use fully-connected layers.

\begin{figure}
\centering
  \subfigure[]{
    \includegraphics[width=1.5in]{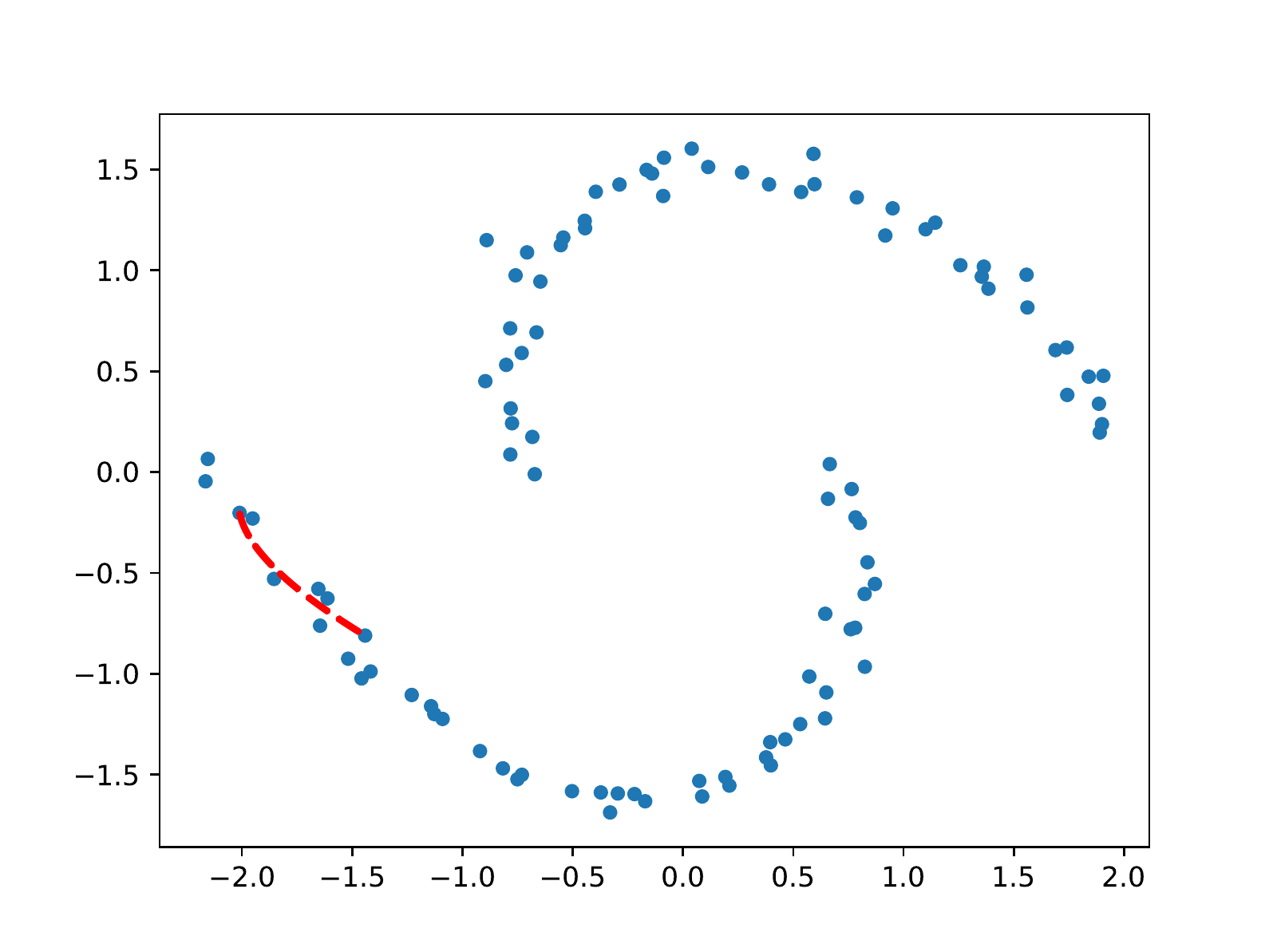}
  }
  \subfigure[]{
    \includegraphics[width=1.5in]{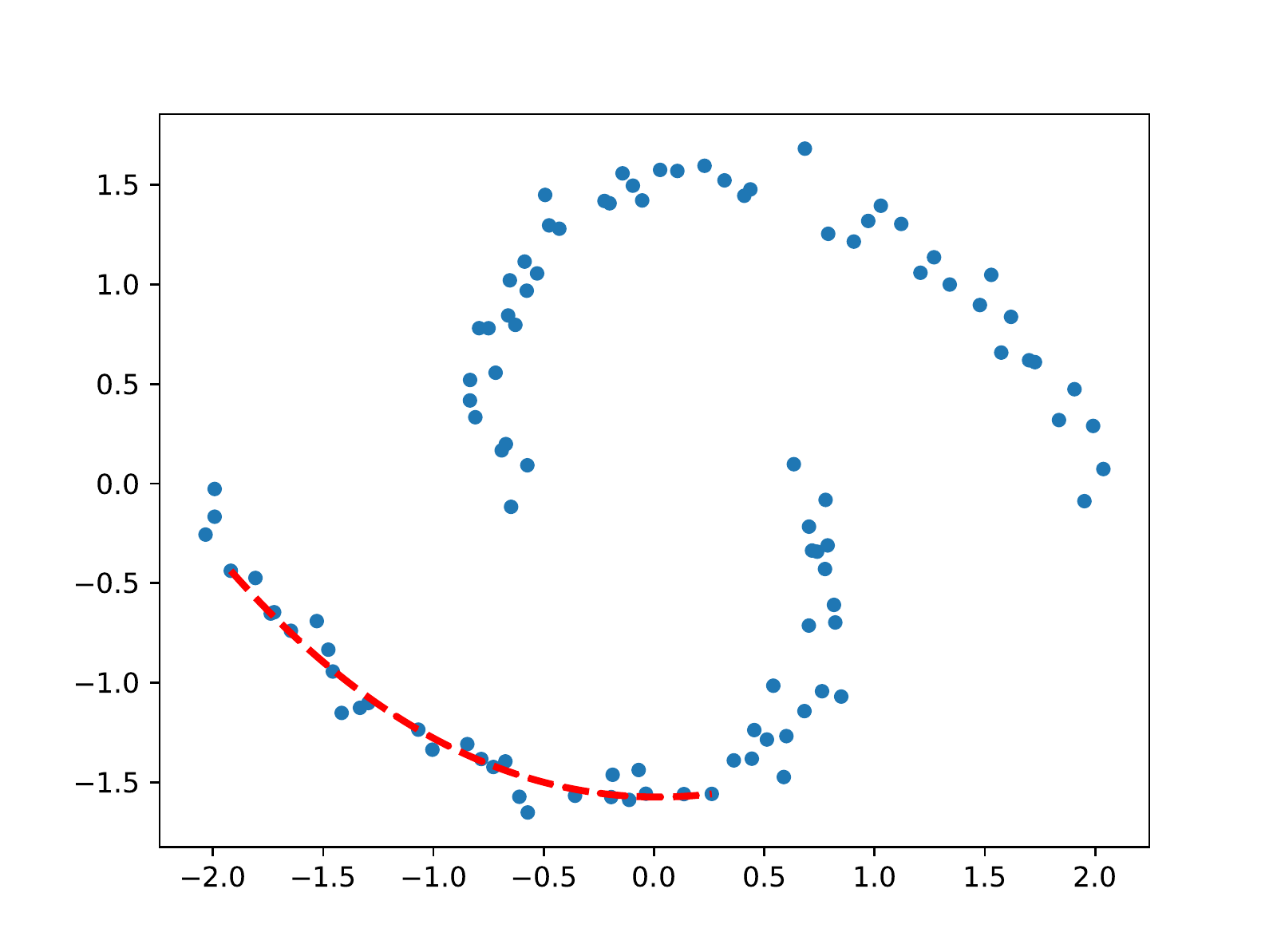}
  }\\
  \subfigure[]{
    \includegraphics[width=1.5in]{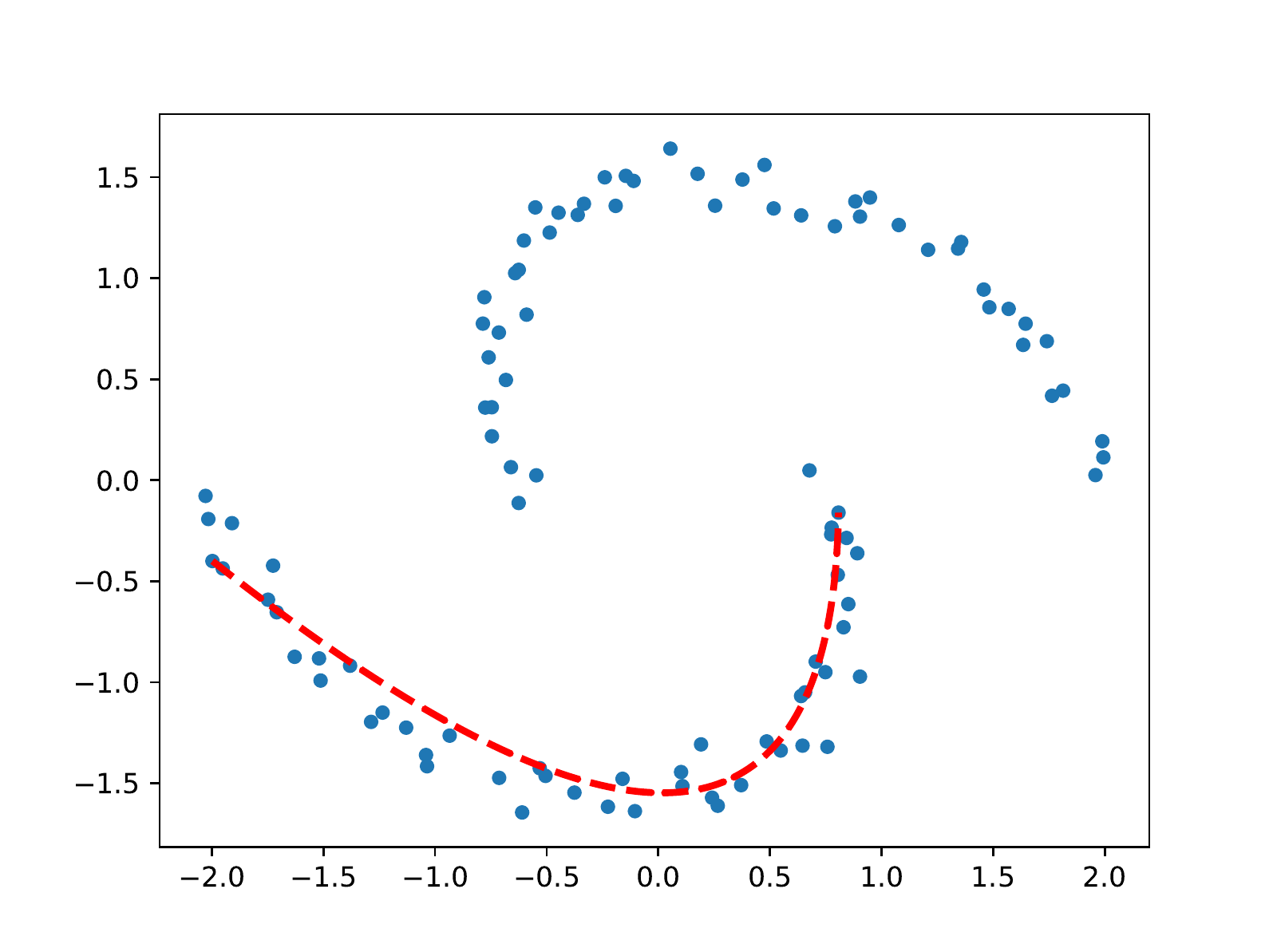}
  }
  \subfigure[]{
    \includegraphics[width=1.5in]{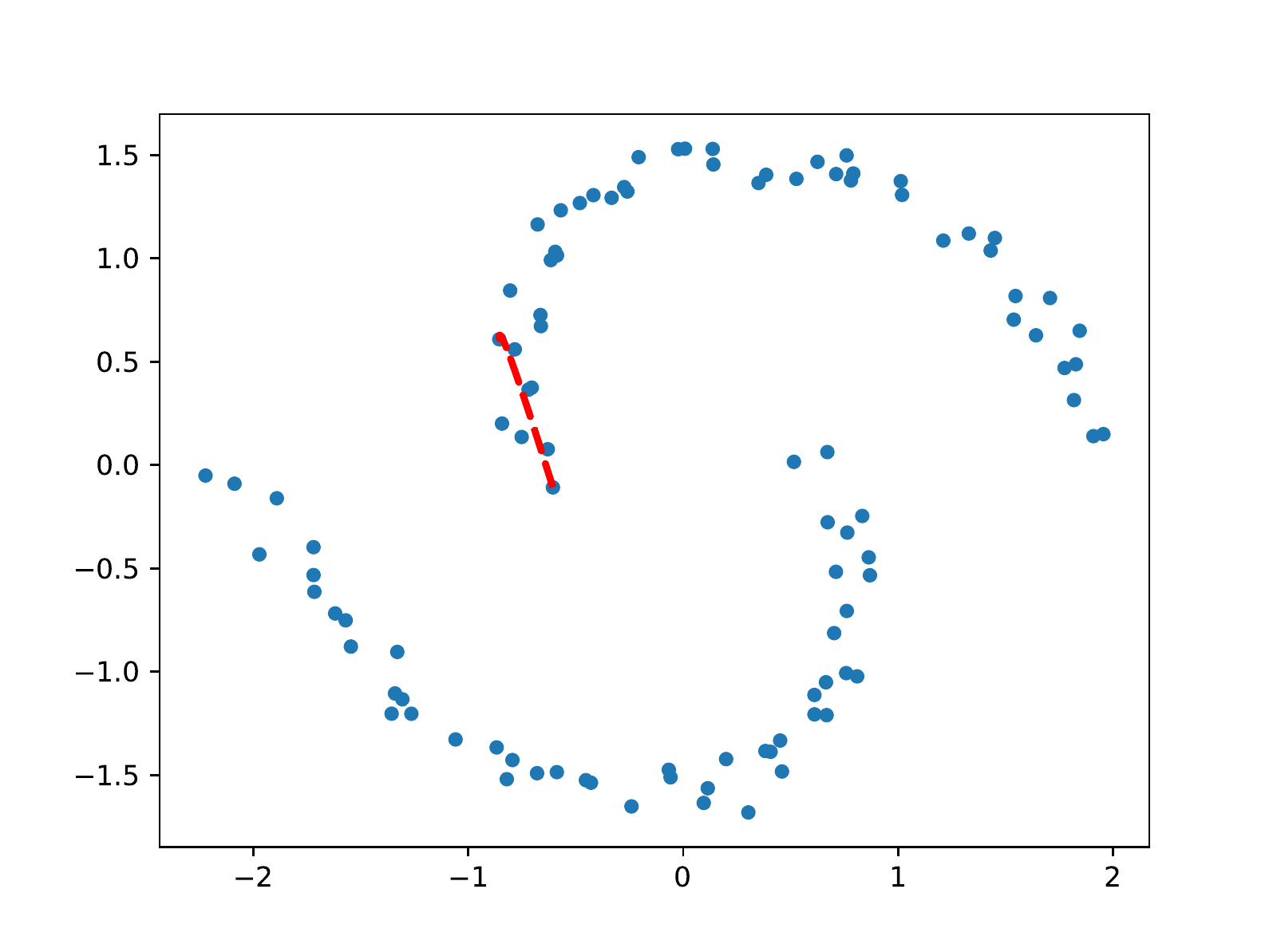}
  }\\
  \subfigure[]{
    \includegraphics[width=1.5in]{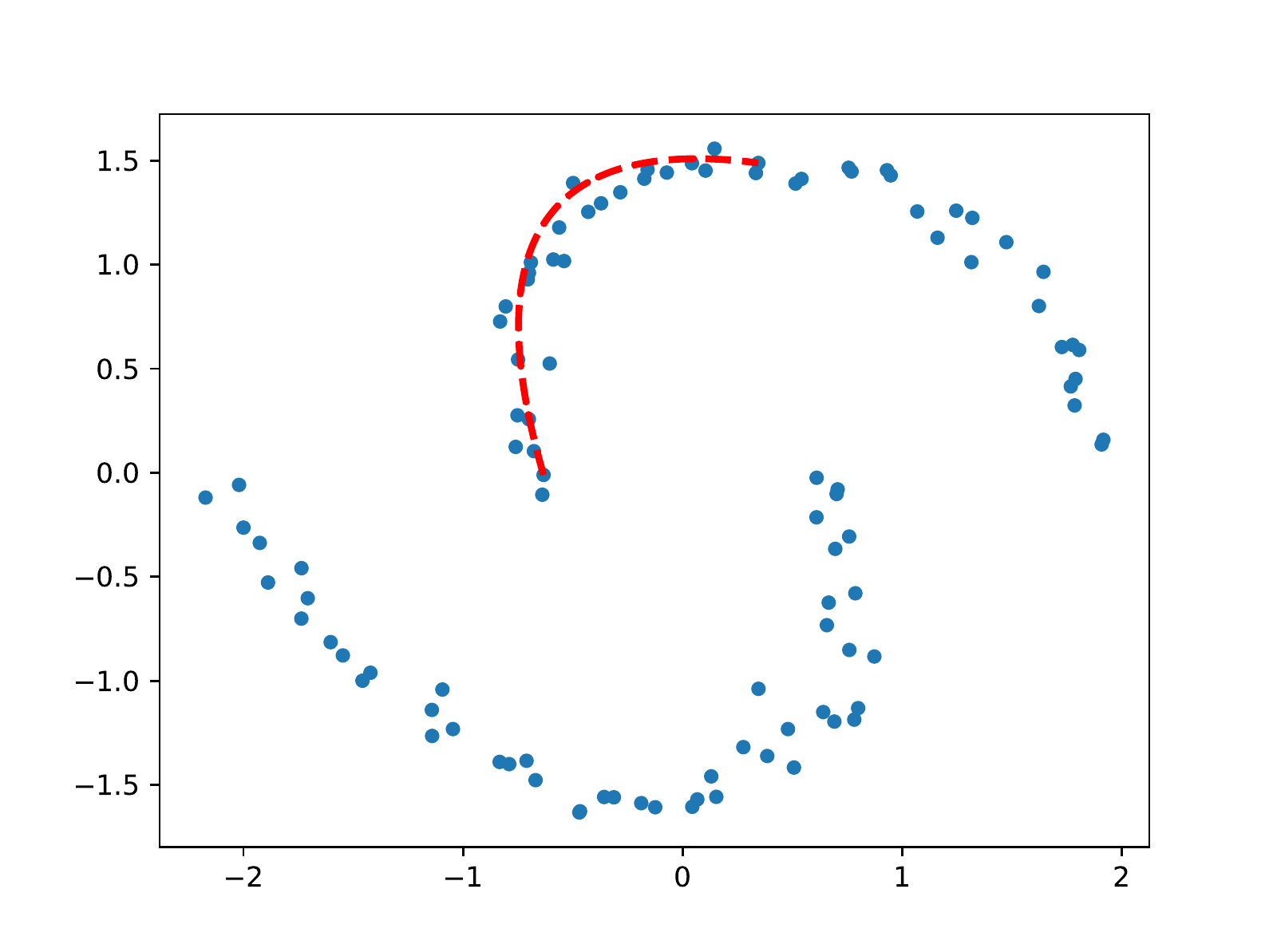}
  }
  \subfigure[]{
    \includegraphics[width=1.5in]{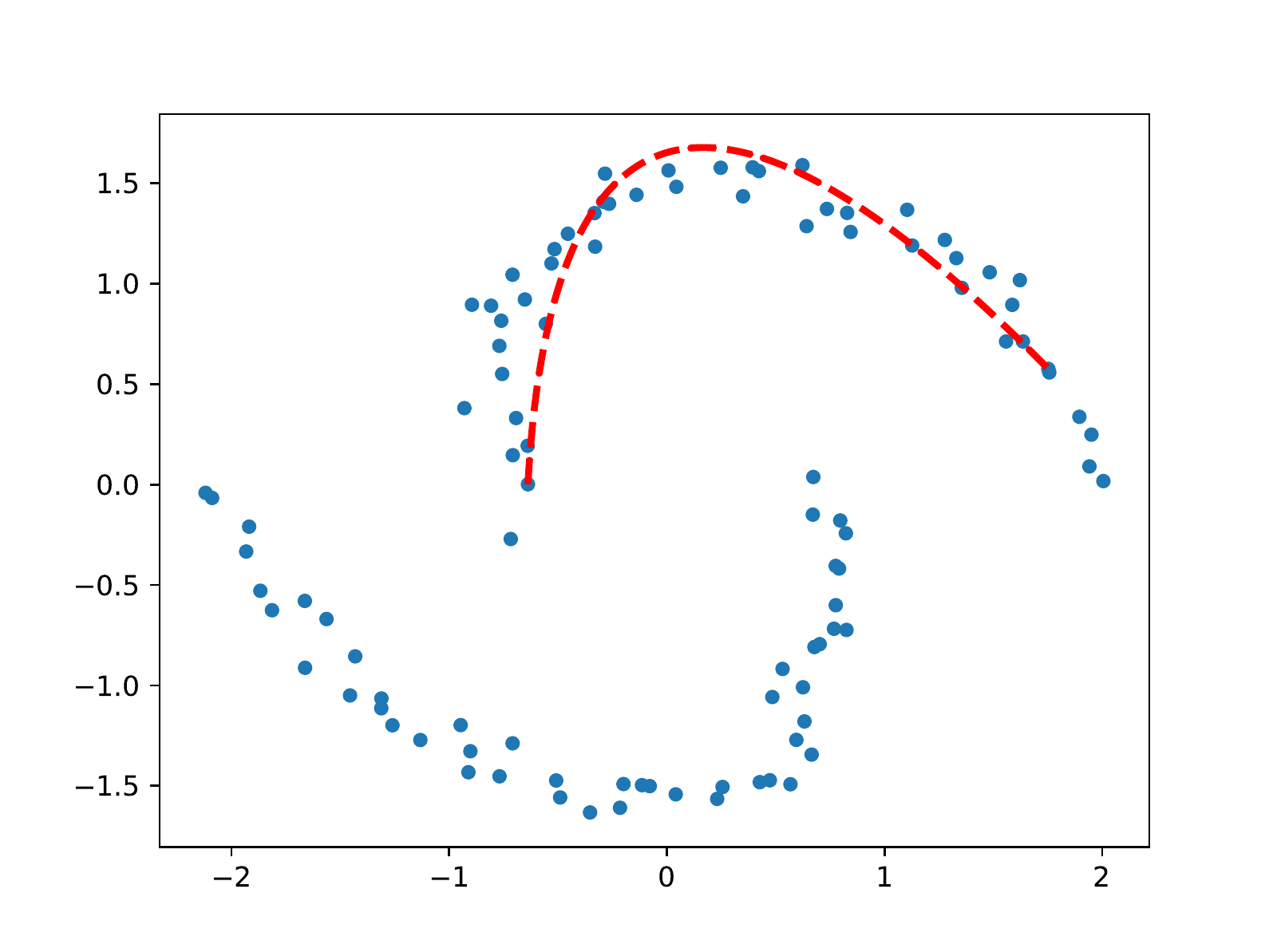}
  }
\caption{Examples of the optimized quadratic curves in latent space for two-moon dataset}
\label{twomoon_geo_exam}
\end{figure}

\begin{figure}
  \centering
  \includegraphics[width=0.5\textwidth]{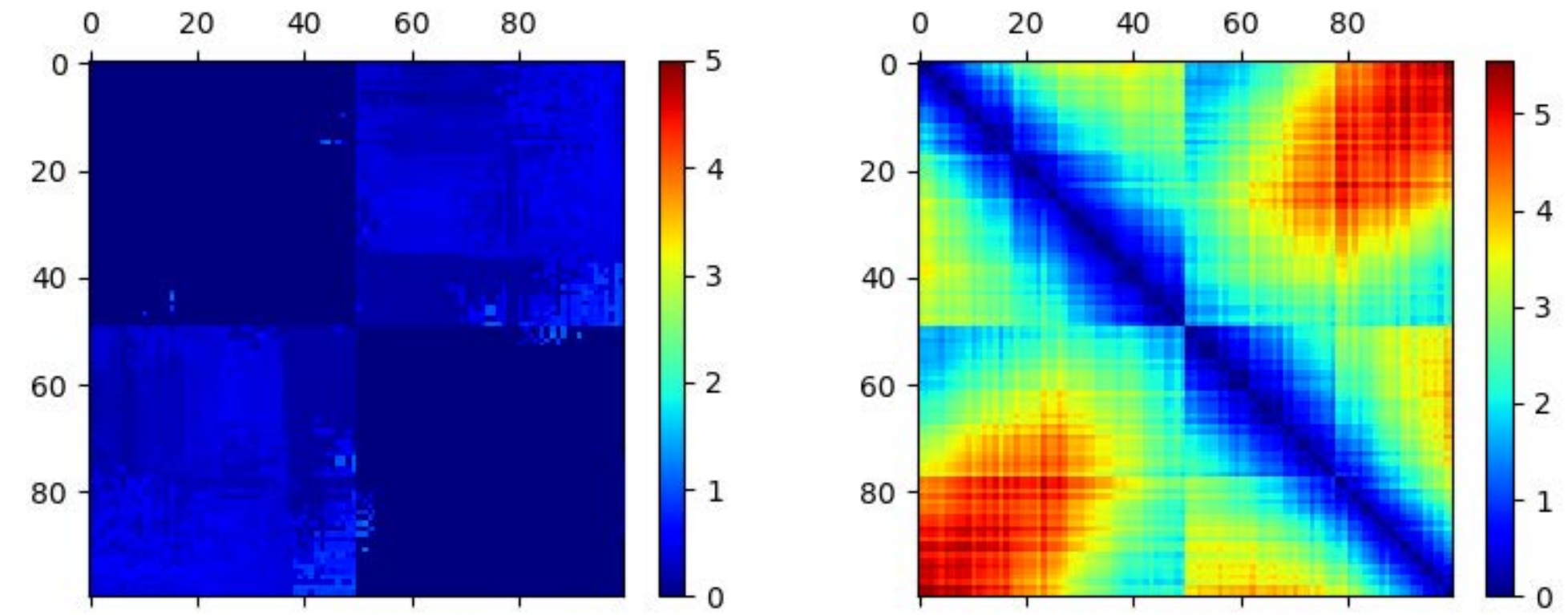}\\
  \caption{The comparison of distance matrix. \textit{Left}: Shortest distance optimized in reconstructed data space. \textit{Right}: Euclidean distance in latent variable space}\label{fig11}
\end{figure}

\begin{figure}
  \centering
  \subfigure[]{
    \includegraphics[width=1.6in]{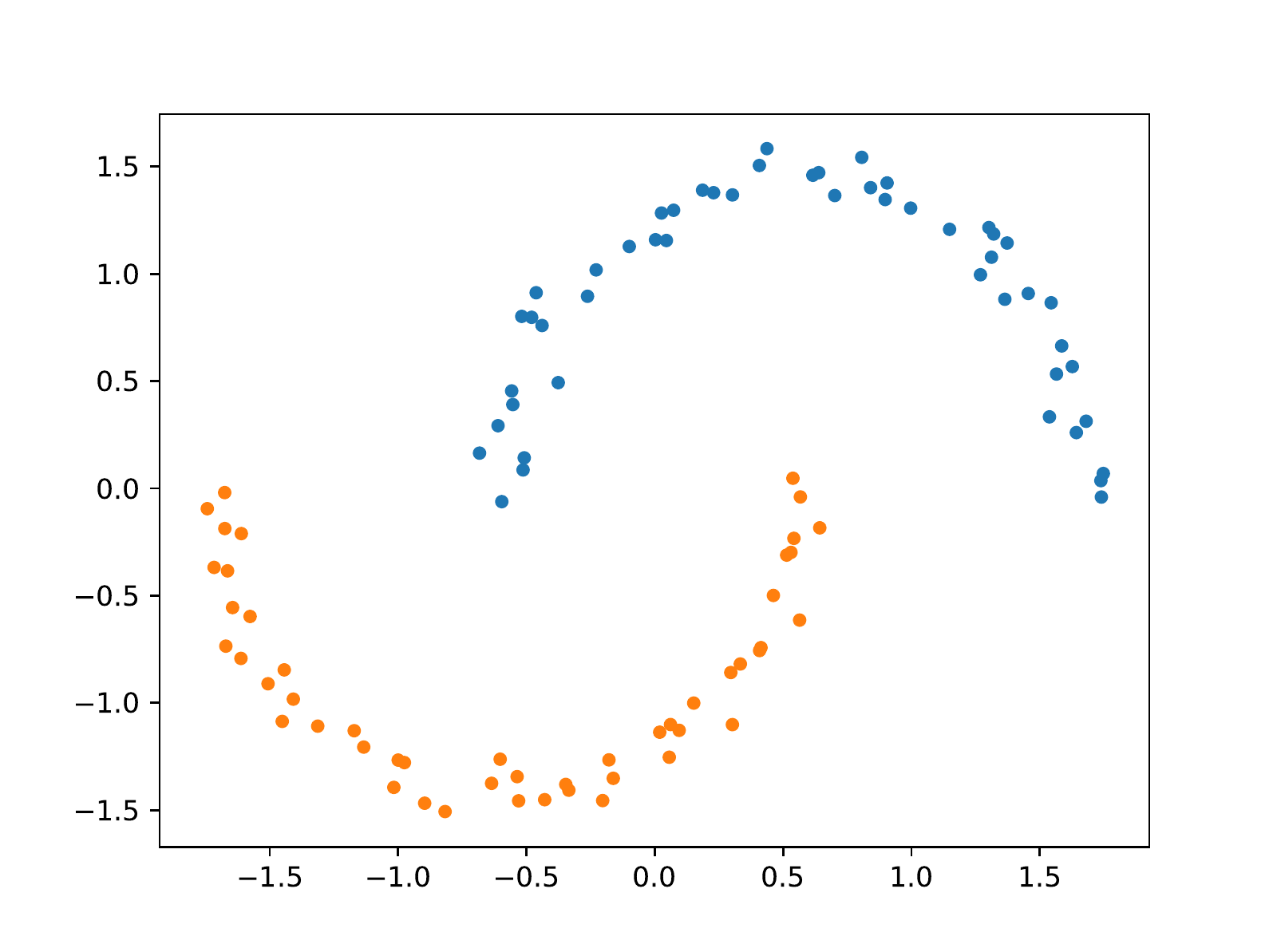}
  }
  \subfigure[]{
    \includegraphics[width=1.6in]{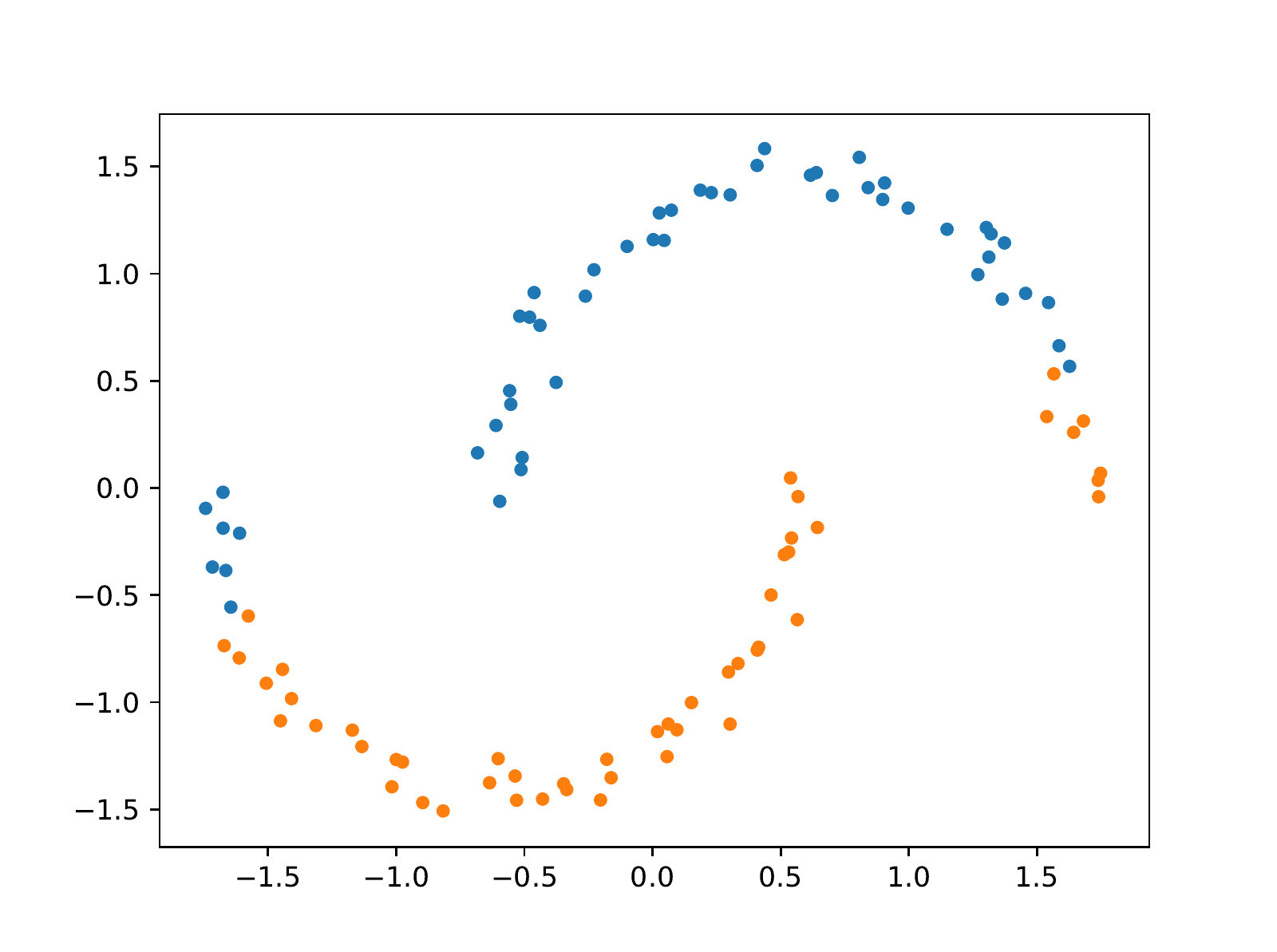}
  }\\
  \subfigure[]{
    \includegraphics[width=1.6in]{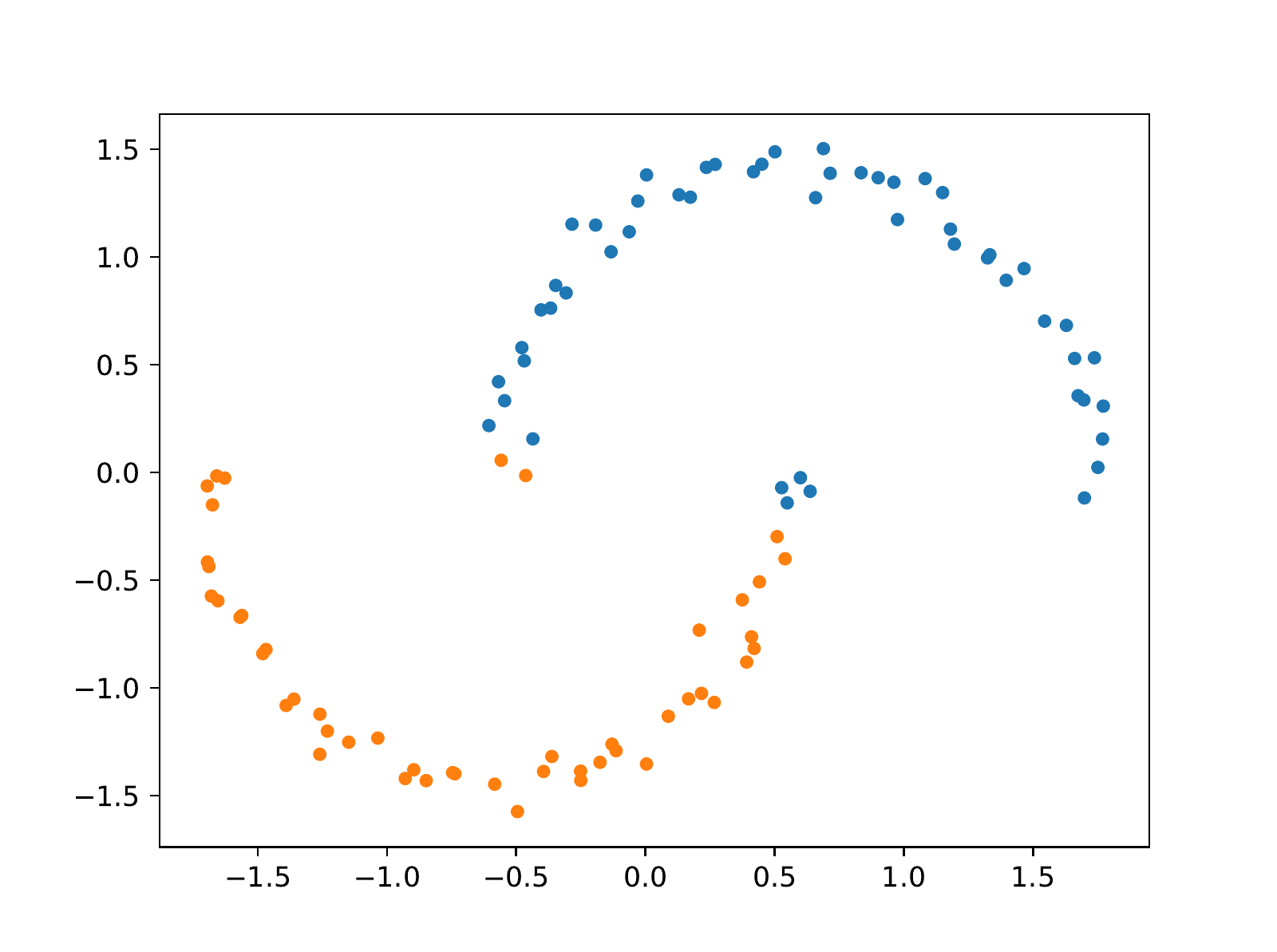}
  }
  \caption{(a): The two-moon clustering result of $k$-medoids with geodesic distance. (b) The result of $k$-medoids with original distance in latent space. (c) The clustering result of \textit{SC}.}
  \label{fig_twomoon} 
\end{figure}

Figure~\ref{twomoon_geo_exam} show the latent space of the resulting VAE along with
several quadratic geodesics. We see that the geodesics nicely follow the structure
of the data. This also influences the observed clustering structure.
Figure~\ref{fig11} show all pairwise distances using both geodesic and Euclidean
distances. I should be noted that the first 50 points belong to the first  ``moon''
while the remaining belong to the other. From the figure, we see that the geodesic
distance reveals the cluster structure much more clearly than the Euclidean counterpart.
We validate this by performing $k$-medoids clustering using the two distances.
As a baseline, we also consider standard \emph{spectral clustering (SC)}
\cite{spectralClus} to the original data.
We report clustering accuracy (the ratio of correct clustered sample number and the number of observations)
in Fig.~\ref{fig_twomoon} and in Table~\ref{twomoon_tab}. It is evident that
the geodesic distance reveals the intrinsic structure of the data.

\begin{table}[!htbp]
\centering
\caption{Two-moon dataset clustering accuracy}\label{twomoon_tab}
\begin{tabular}{cccc}
\toprule
method & $k$-medoids & $k$-medoids & \textit{SC}\\
\midrule
data samples & reconstructed data & latent variable & original data\\
distance & Geodesic & Euclidean & Euclidean\\
accuracy & 1 & 0.86 & 0.92\\
\bottomrule
\end{tabular}
\end{table}

\subsubsection{Synthetic Anisotropically Distributed Data}
\begin{figure*}
  \centering
  \subfigure[]{
    \includegraphics[width=2.0in]{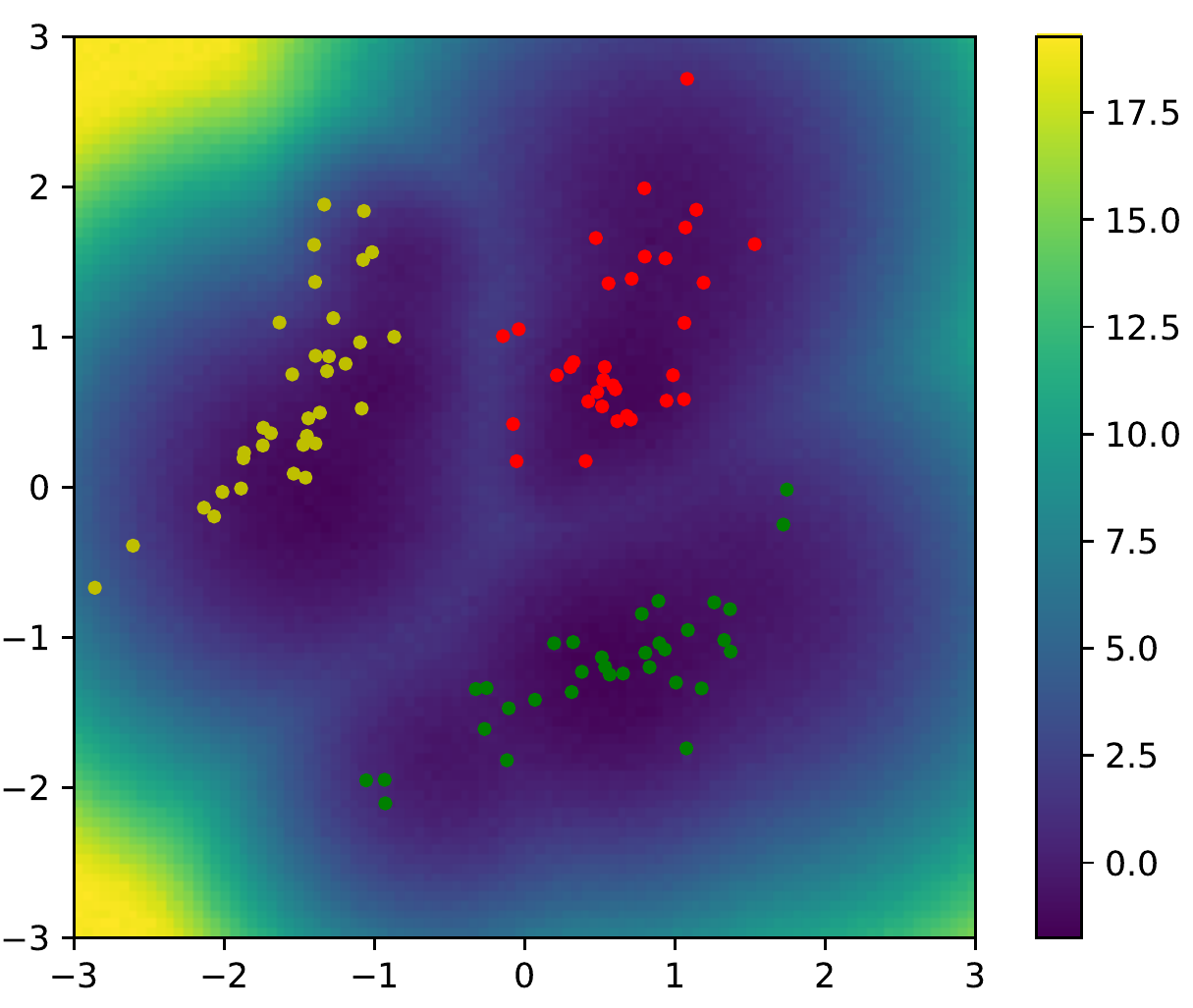}
  }
  \subfigure[]{
    \includegraphics[width=4.4in]{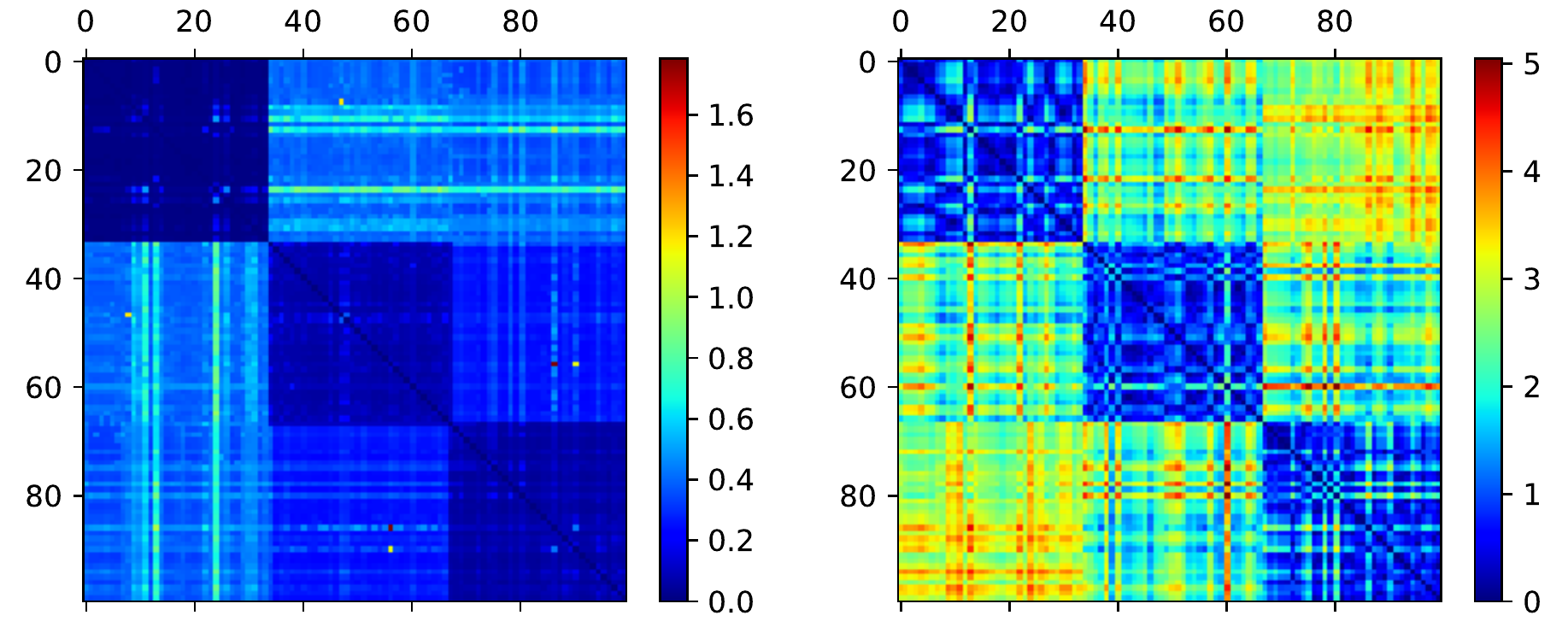}
  }
  \caption{(a): The logarithm of the volume measure in the latent space. (b) Left: The optimized geodesic pair-wise distance. Right: The Euclidean pair-wise distance in latent space.}
  \label{fig_aniso} 
\end{figure*}

\begin{table}[!htbp]
\centering
\caption{Anisotropic data samples clustering accuracy}\label{aniso_tab}
\begin{tabular}{cccc}
\toprule
method & $k$-medoids & $k$-medoids & \textit{SC}\\
\midrule
data samples & reconstructed data & latent variable & original data\\
distance & Geodesic & Euclidean & Euclidean\\
accuracy & 1 & 0.80 & 0.96\\
\bottomrule
\end{tabular}
\end{table}

\begin{figure}
  \centering
  \subfigure[]{
    \includegraphics[width=1.6in]{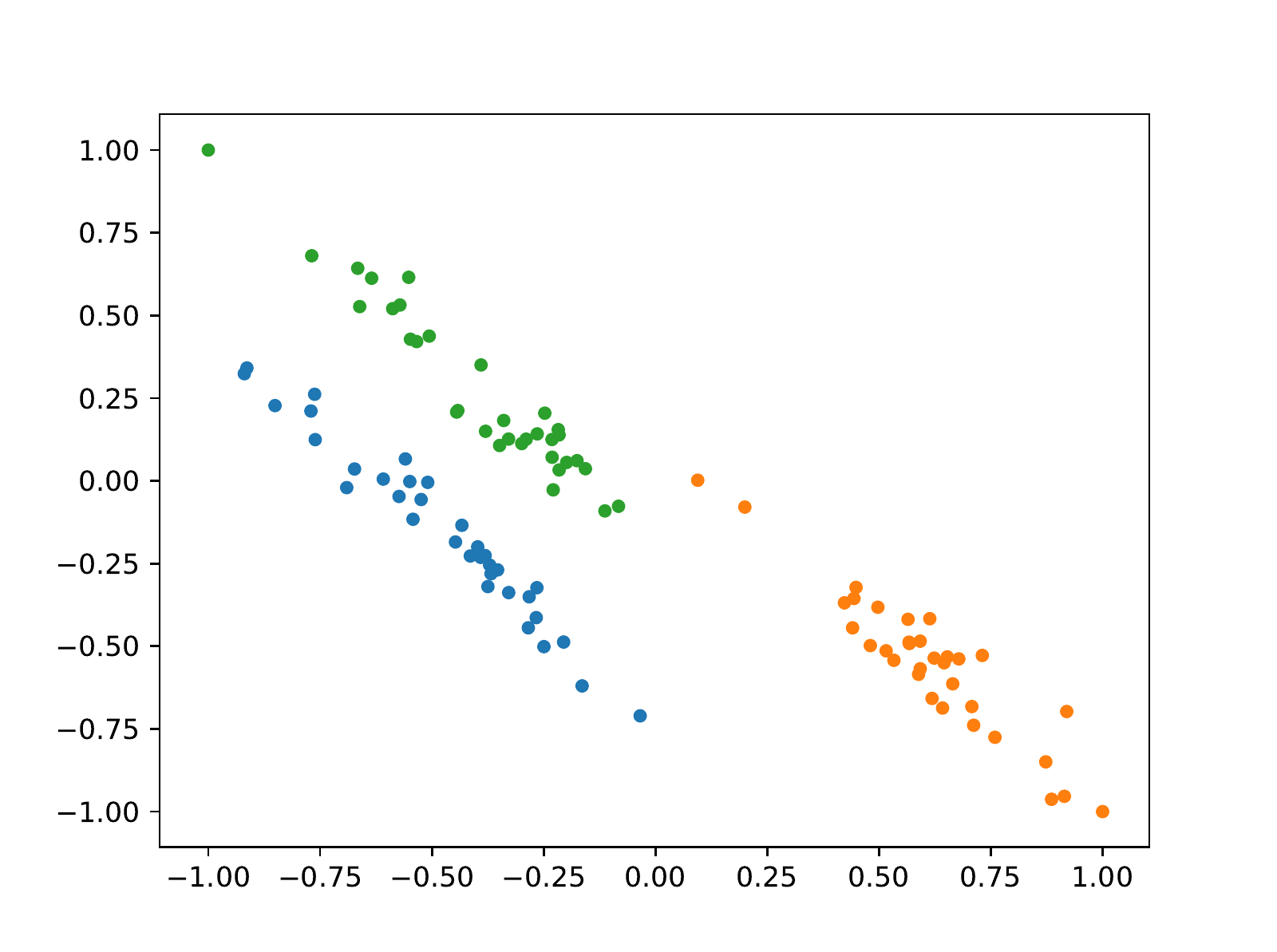}
  }
  \subfigure[]{
    \includegraphics[width=1.6in]{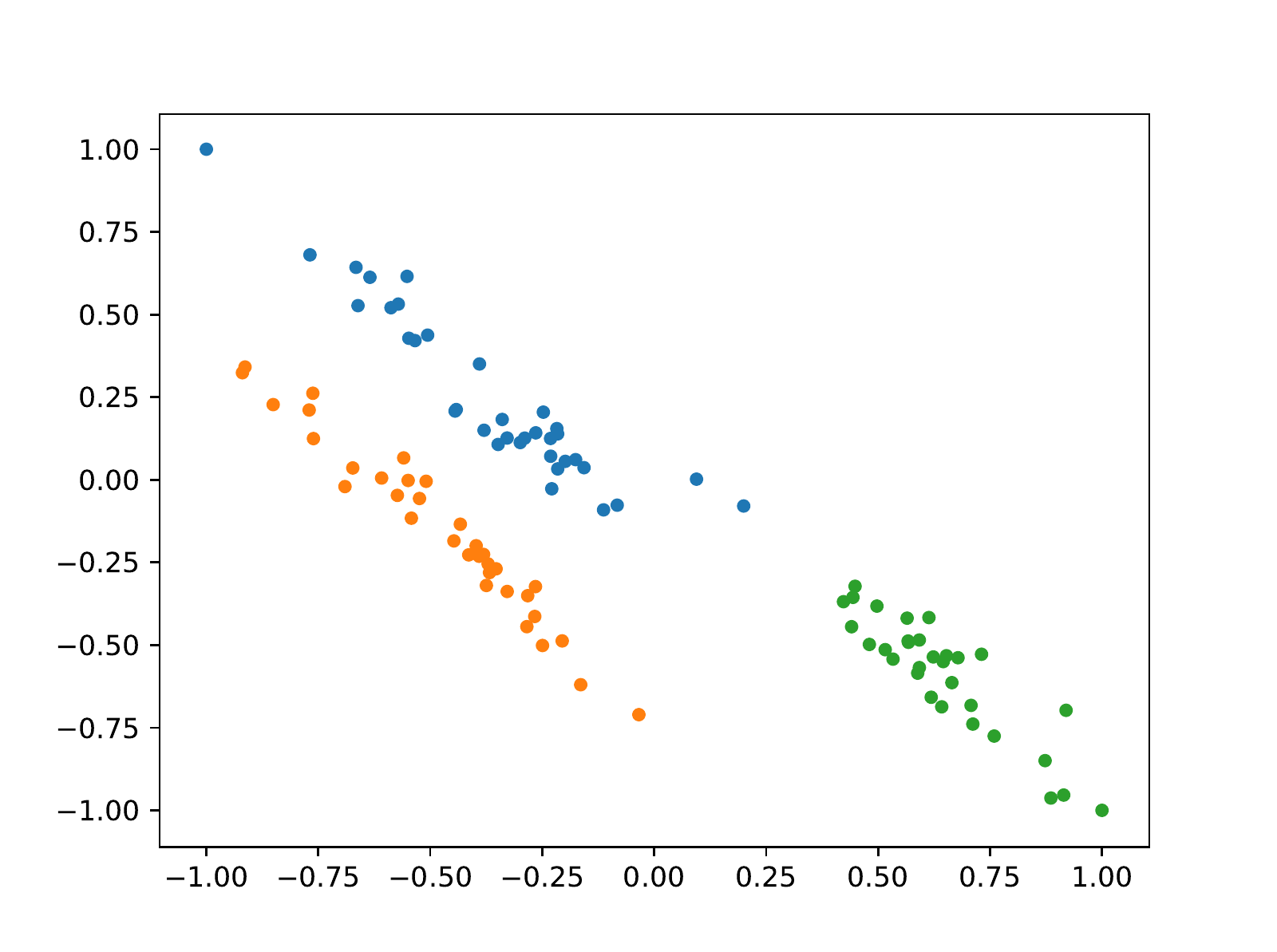}
  }\\
  \subfigure[]{
    \includegraphics[width=1.6in]{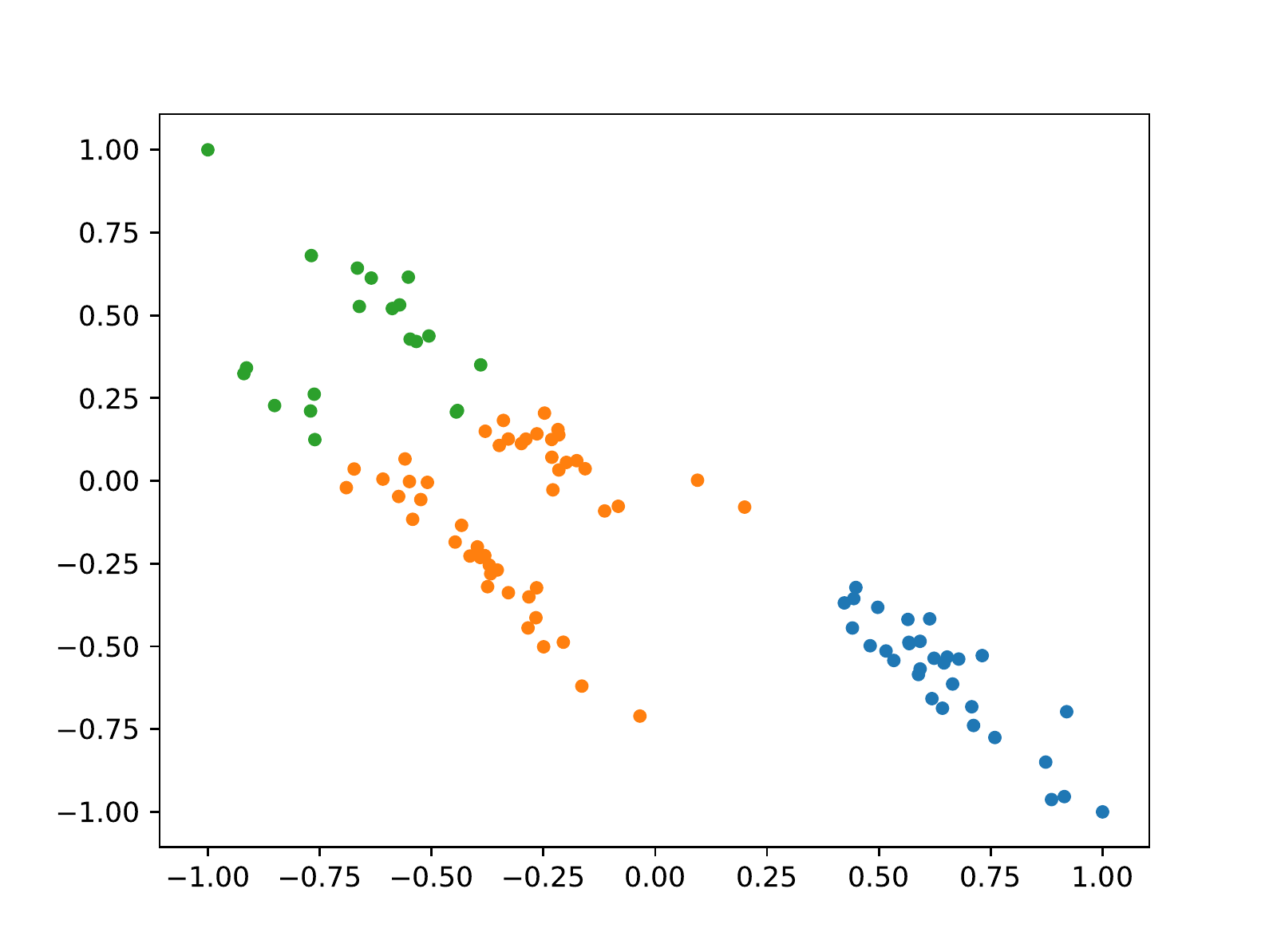}
  }
  \caption{(a): The anisotropic scattered samples clustering result of $k$-medoids with geodesic distance. (b) The result of $k$-medoids with original distance in latent space. (c) The clustering result of \textit{SC}.}
  \label{aniso_figures} 
\end{figure}

Using the same setup as for the two-moon dataset, we generate 100 samples from
clusters with anisotropic distributions.
Figure~\ref{fig_aniso} shows both volume measure and pair-wise distances.
Again $k$-medoids clustering show that the geodesic distance does a much better
job at capturing the data structure than the baselines. Clustering accuracy
is in Table~\ref{aniso_tab} and the found clusters are shown in Fig.~\ref{aniso_figures}.

\subsubsection{The MNIST Dataset}
\begin{figure}
  \centering
  \subfigure[]{
    \includegraphics[width=1.4in]{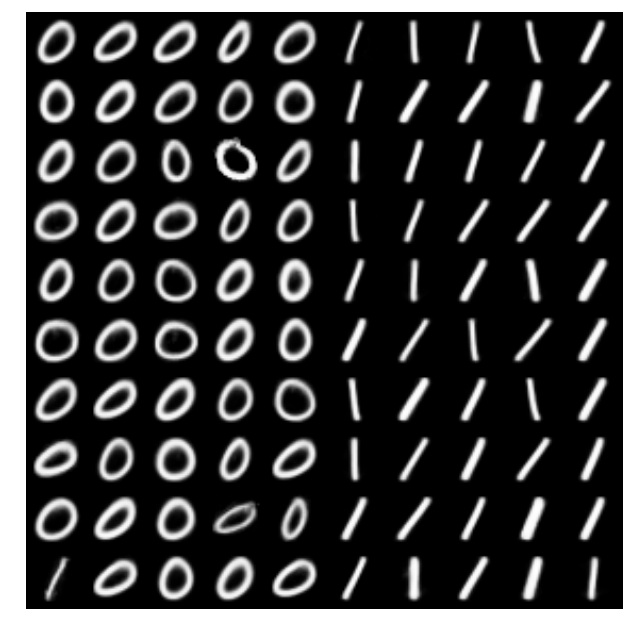}
  }
  \subfigure[]{
    \includegraphics[width=1.4in]{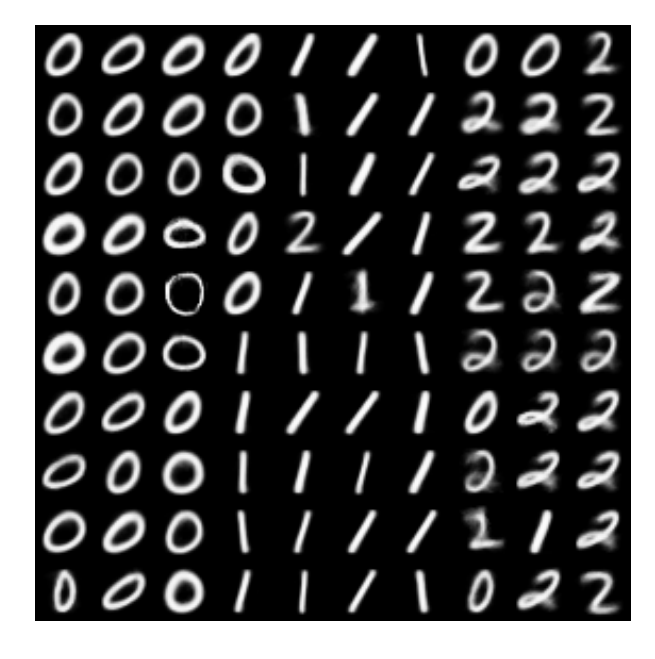}
  }
  \caption{Generated '0', '1' and '2' examples for MNIST dataset by the VAE used in this paper.}
  \label{mnist_draw} 
\end{figure}

From the well-known MNIST dataset, we take hand-written digit '0', '1' and '2'
to test 2-class and 3-class clustering. For \textit{H-enc} and \textit{H-dec}
layers, we use two hidden fully-connected layers with Relu activations\footnote{The
  number of \textit{H-enc} neural nodes: from 784 to 500 and from 500 to 2.
  The number of \textit{H-dec} neural nodes: from 2 to 500 and from 500 to 784.},
and for the \textit{S-enc} layer, we use one fully-connected layer with a sigmoid
activation function, and for \textit{M-enc}, \textit{M-dec} layers we use
fully-connected layers with \textit{identity} activation functions.
Images generated by both networks are shown in Fig.~\ref{mnist_draw}.

\begin{figure*}
  \centering
  \subfigure[]{
    \includegraphics[width=2.0in]{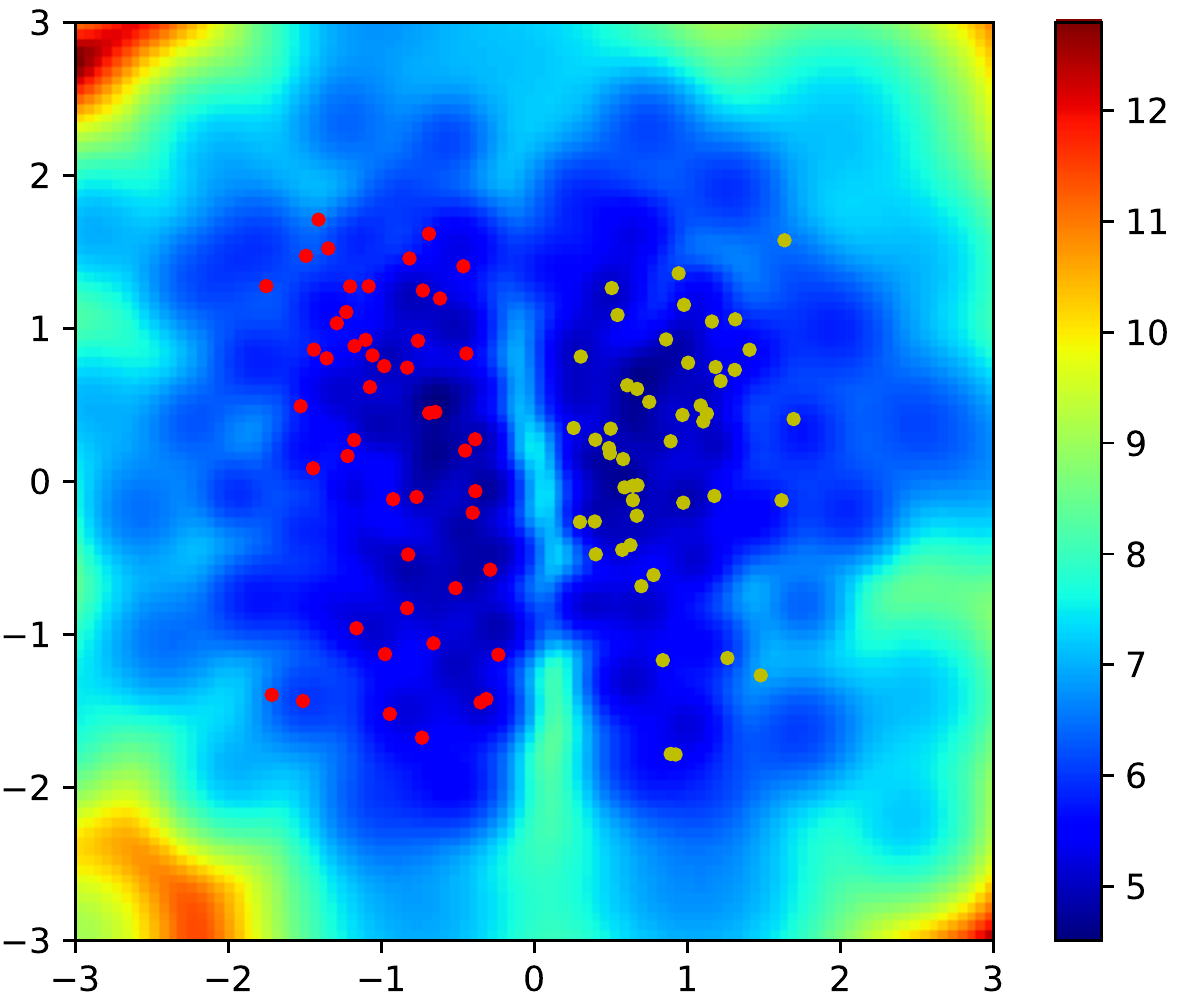}
  }
  \subfigure[]{
    \includegraphics[width=4.4in]{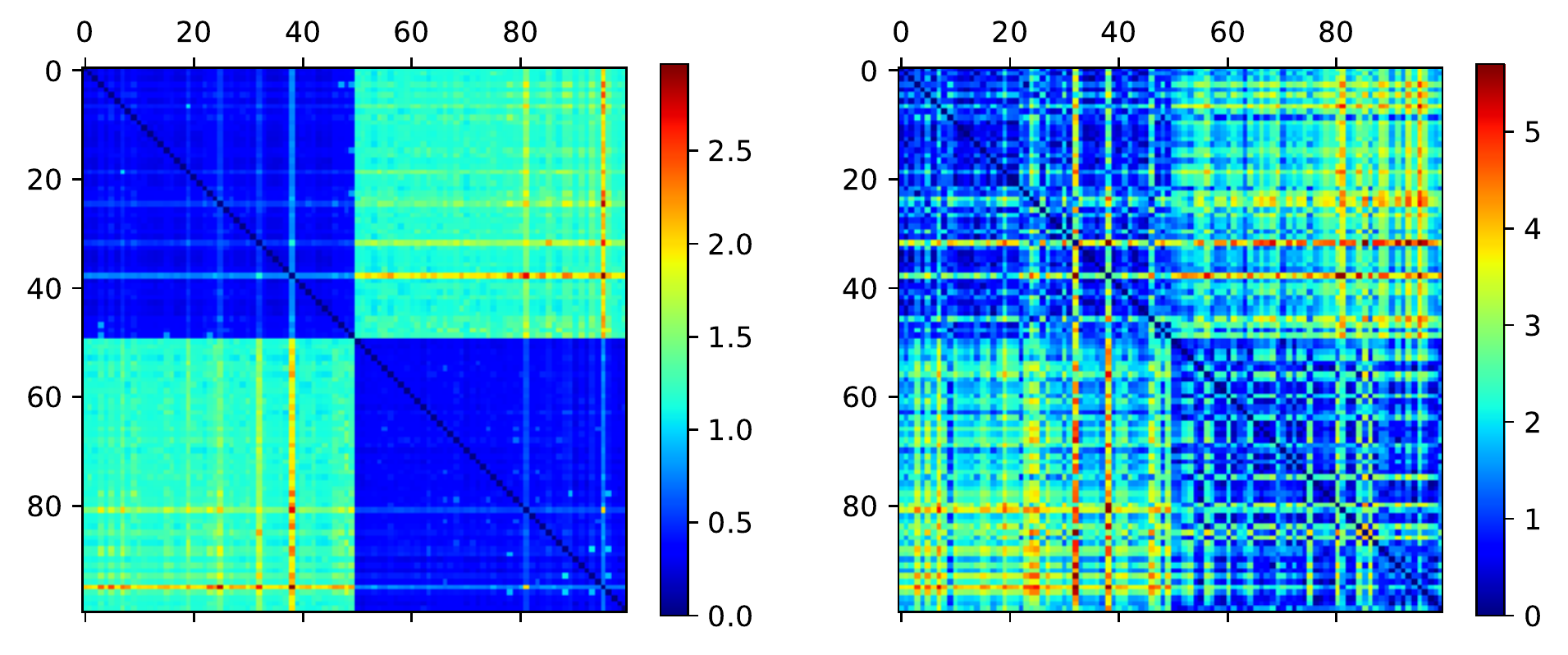}
  }
  \caption{(a): The logarithm of the volume measure in latent space for '0' and '1' digit images. (b) Left: pair-wise geodesic distances. Right: pair-wise Euclidean distances in latent space.}
  \label{m01_dist} 
\end{figure*}

\begin{figure*}
  \centering
  \subfigure[]{
    \includegraphics[width=2.1in]{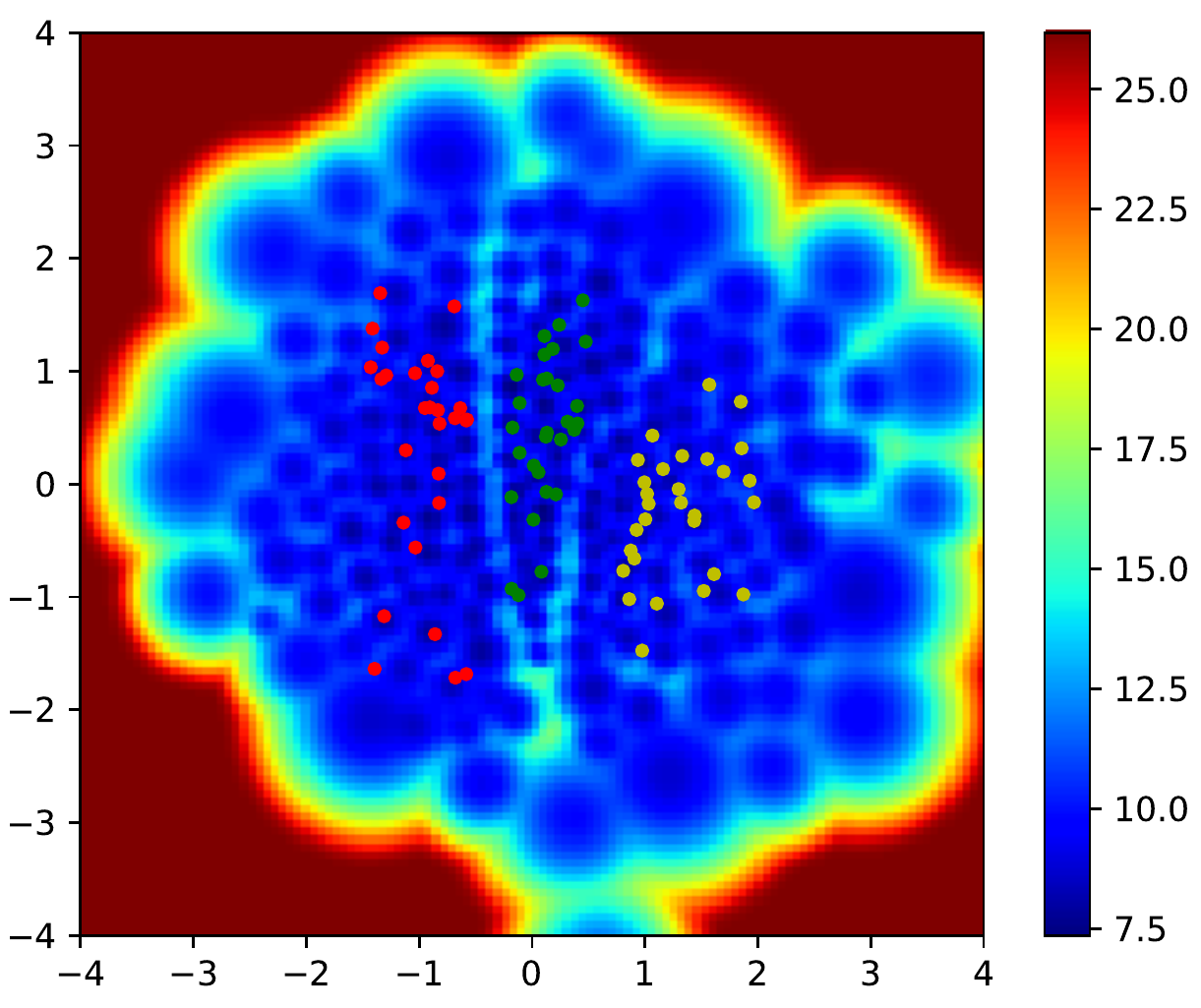}
  }
  \subfigure[]{
    \includegraphics[width=4.4in]{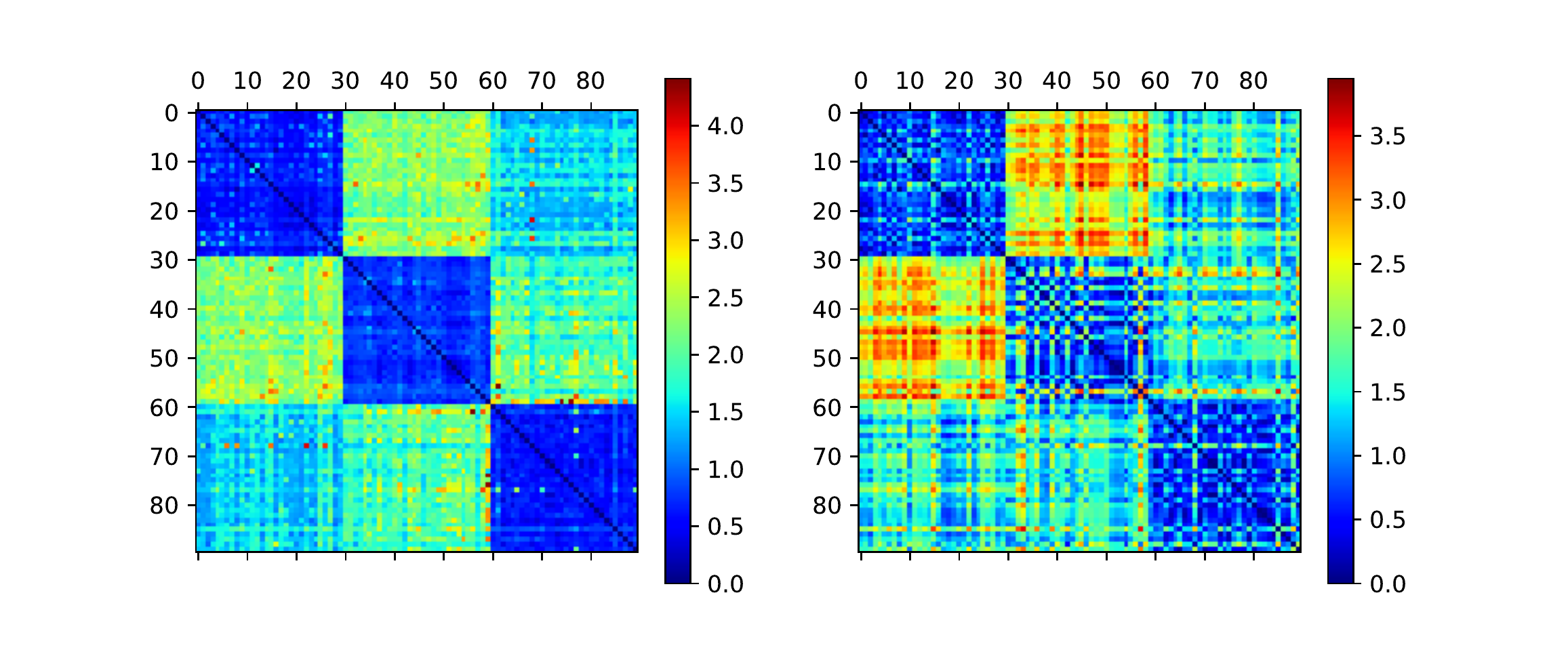}
  }
  \caption{(a): The logarithm of the volume measure in latent space for '0', '1' and '2' digit images. (b) Left: The optimized geodesic pair-wise distance. Right: The Euclidean pair-wise distance in latent space.}
  \label{m012_dist} 
\end{figure*}

For the 2-class situation, we use digits '0' and '1'.
We select 50 samples from each class and compute their pair-wise distances,
which are shown in Fig.~\ref{m01_dist}.
For the 3-class situation, we select 30 samples from each class and show pair-wise
distances in Fig.~\ref{m012_dist}.
In both cases, the geodesic distance reveals a clear clustering structure.
We also see this in $k$-medoids clustering, which outperforms the baselines
(Table~\ref{tab2}).

\begin{table}[!htbp]
\centering
\caption{MNIST dataset clustering 2-class and 3-class accuracy}\label{tab2}
\begin{tabular}{cccc}
\toprule
method & $k$-medoids & $k$-medoids & \textit{SC}\\
\midrule
data samples & reconstructed data & latent variable & original data\\
distance & Geodesic & Euclidean & Euclidean\\
'0'-'1' accuracy & 1 & 0.93 & 0.69\\
'0'-'1'-'2' accuracy & 1 & 0.80 & 0.36\\
\bottomrule
\end{tabular}
\end{table}

\subsubsection{The Fashion-MNIST Dataset}

\begin{figure}
  \centering
  \subfigure[]{
    \includegraphics[width=1.4in]{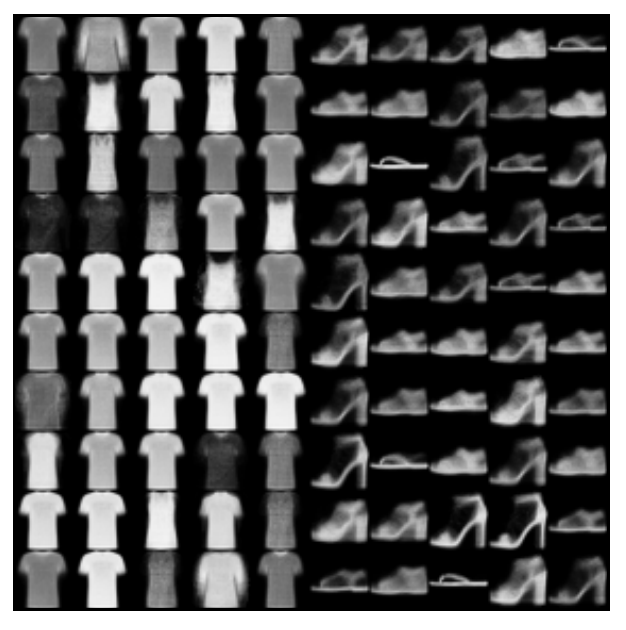}
  }
  \subfigure[]{
    \includegraphics[width=1.4in]{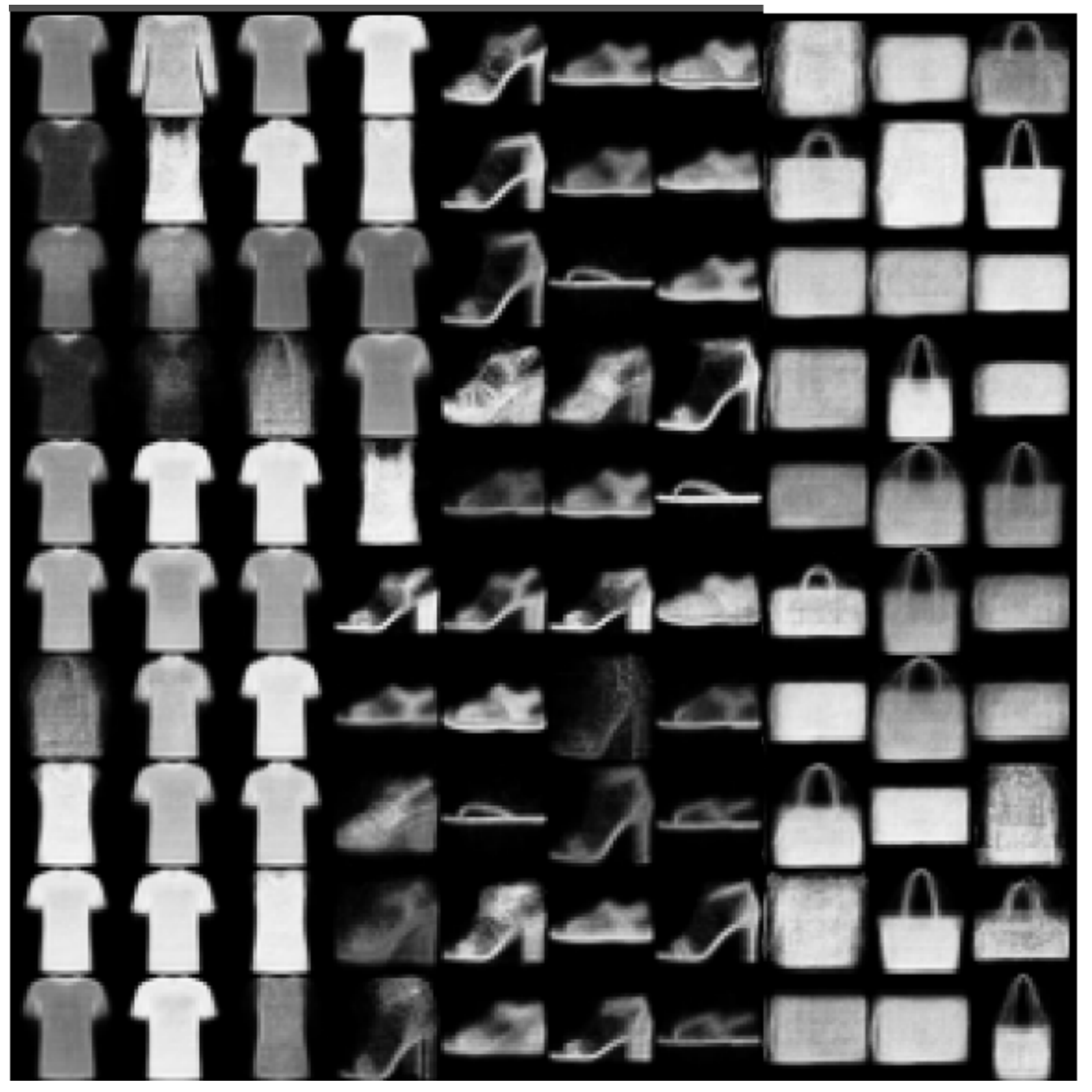}
  }
  \caption{Generated 'T-shirt', 'Sandal' and 'Bag' examples for Fashion-MNIST dataset by VAE used in this paper.}
  \label{fmnist_draw} 
\end{figure}

Fashion-MNIST \cite{xiao2017/online} is a dataset of Zalando's article images. Each image is
a $28\times 28$ gray-scale image. We consider classes 'T-shirt', 'Sandal' and 'Bag'
to test 2-class and 3-class clustering. For \textit{H-enc} and \textit{H-dec} layers,
we use three hidden fully-connected layers with Relu activations\footnote{The number
  of \textit{H-enc} neural nodes: from 784 to 500 and from 500 to 200 and from 200
  to 100. The number of \textit{H-dec} neural nodes: from 100 to 200 and from
  200 to 500 and from 500 to 784}, and for \textit{S-enc} layer, we use one
fully-connected layer with a sigmoid activation function, and for
\textit{M-enc}, \textit{M-dec} we use fully-connected layers with identity
and sigmoid activation functions respectively.
Images generated by the networks are shown in Fig.~\ref{fmnist_draw}.

\begin{figure*}
  \centering
  \subfigure[]{
    \includegraphics[width=2.1in]{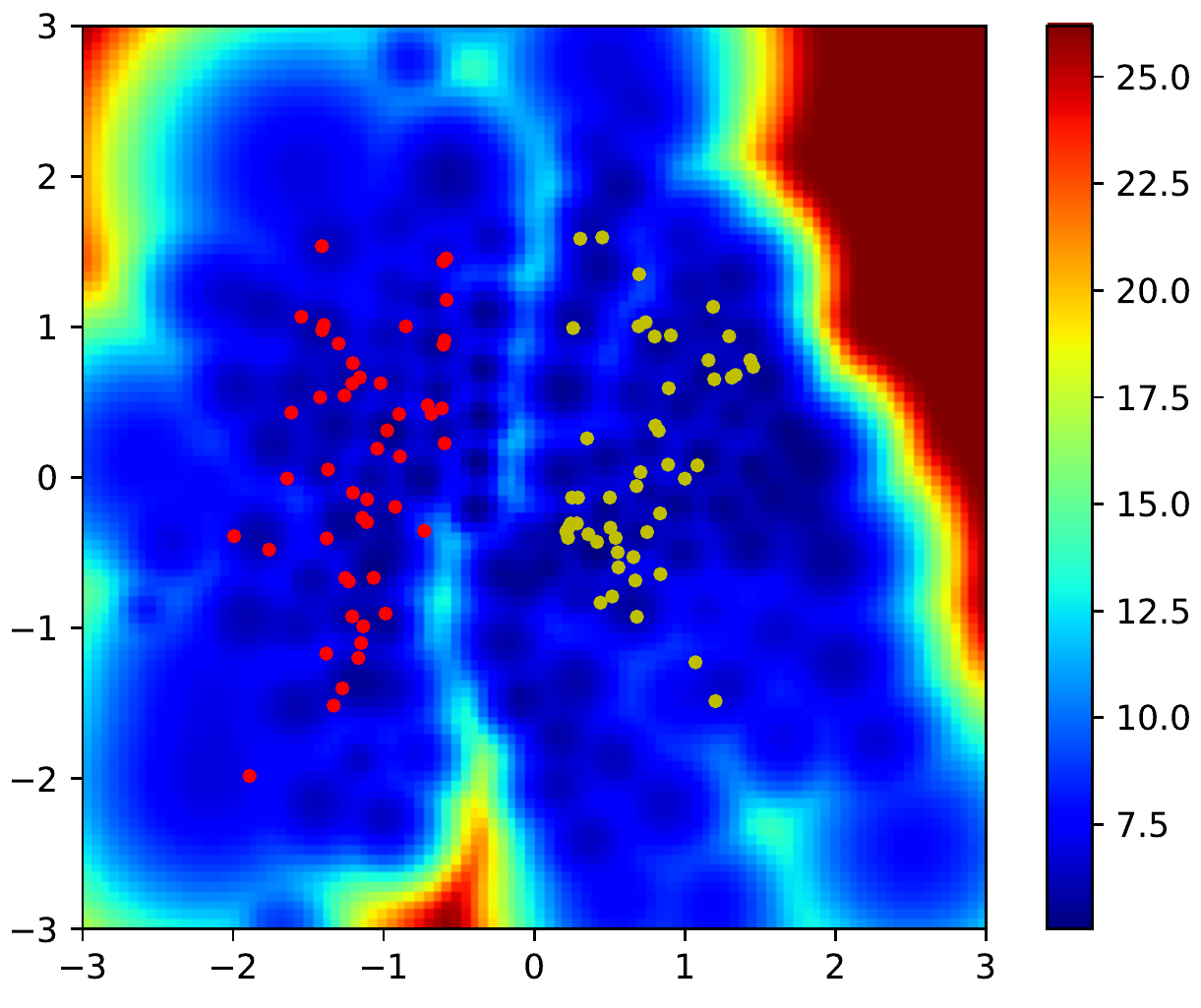}
  }
  \subfigure[]{
    \includegraphics[width=4.4in]{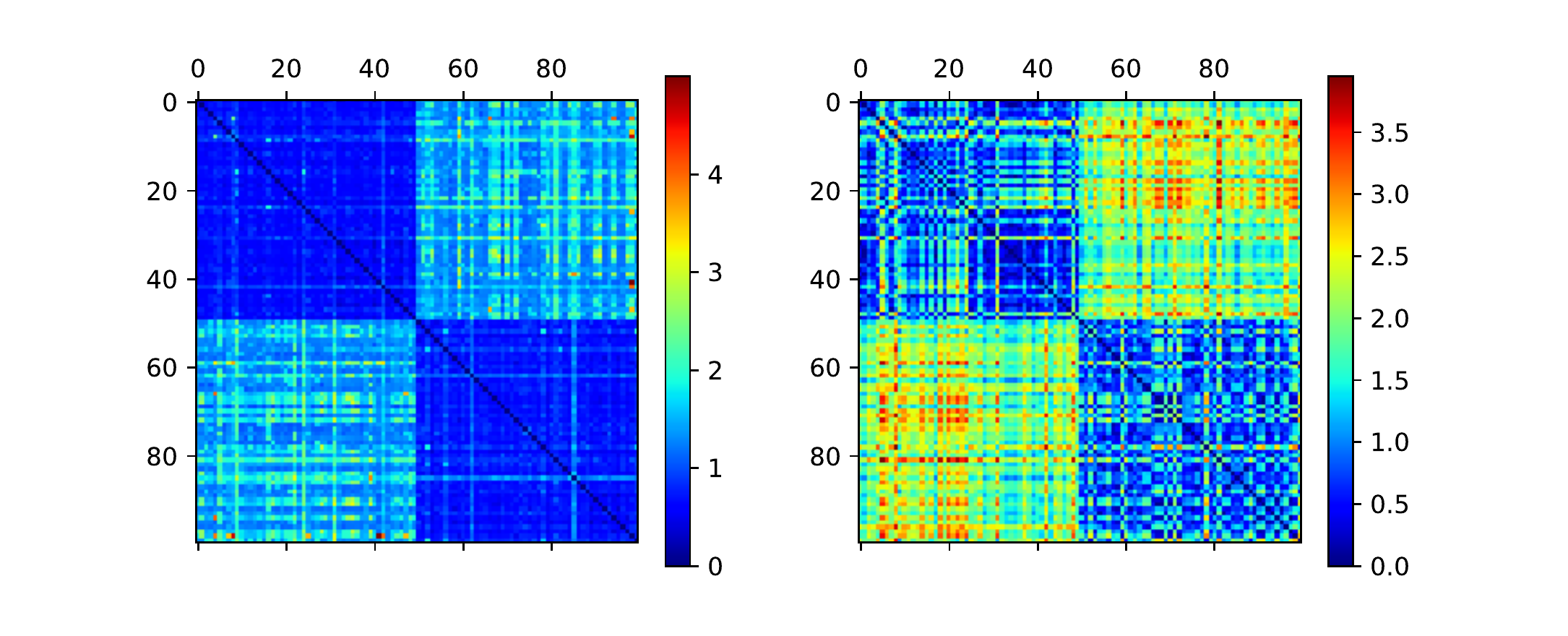}
  }
  \caption{(a): The logarithm of the volume measure in latent space for 'T-shirt' and 'Sandal' images. (b) Left: The optimized geodesic pair-wise distance. Right: The Euclidean pair-wise distance in latent space.}
  \label{f05_dist} 
\end{figure*}

\begin{figure*}
  \centering
  \subfigure[]{
    \includegraphics[width=2.1in]{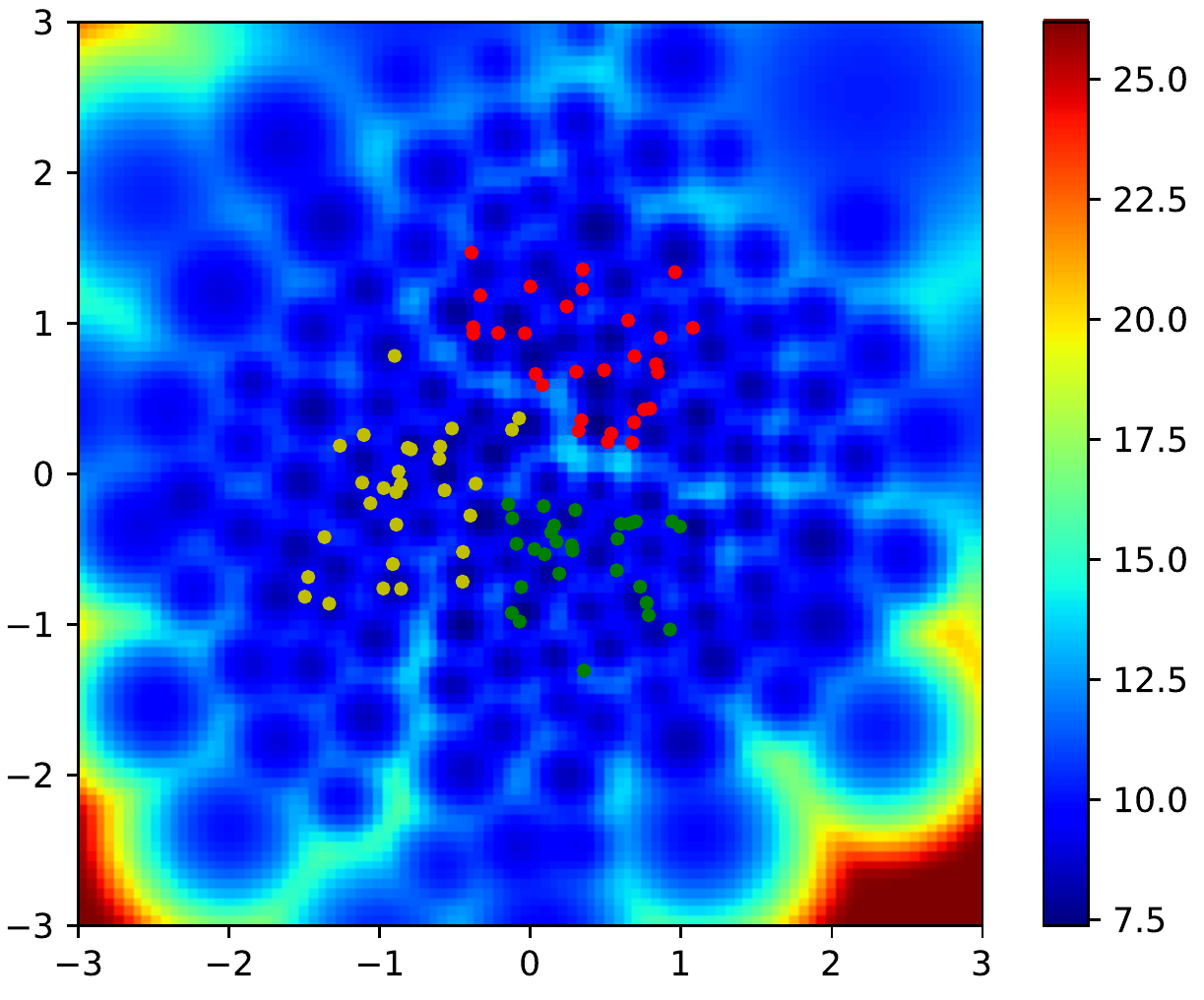}
  }
  \subfigure[]{
    \includegraphics[width=4.4in]{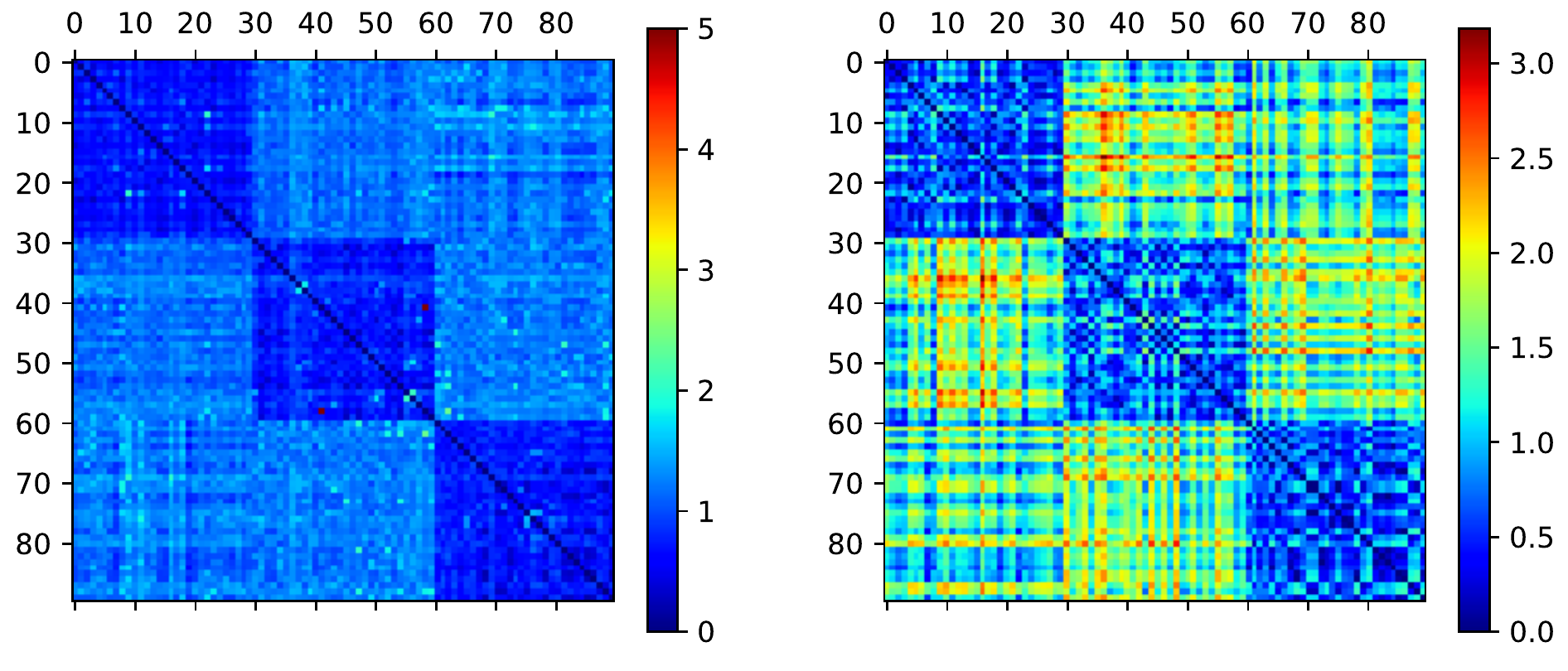}
  }
  \caption{(a): The logarithm of the volume measure in latent space for 'T-shirt' ,'Sandal' and 'Bag' images. (b) Left: The optimized geodesic pair-wise distance. Right: The Euclidean pair-wise distance in latent space.}
  \label{f051_dist} 
\end{figure*}

For the 2-class situation, we use the 'T-shirt' and 'Sandal' samples to train the
VAE. We select 50 samples from 'T-shirt' and 'Sandal' dataset respectively, and compute
pair-wise distances (see Fig.~\ref{f05_dist}).
For the 3-class situation we select 30 samples from each class and compute distances (Fig.~\ref{f051_dist}).
As before, we see that $k$-medoids clustering with geodesic distances significantly
outperform the baselines; see Table~\ref{tab3} for numbers.

\begin{table}[!htbp]
\centering
\caption{Fashion-MNIST dataset clustering 2-class and 3-class accuracy}\label{tab3}
\begin{tabular}{cccc}
\toprule
method & $k$-medoids & $k$-medoids & \textit{SC}\\
\midrule
data samples & generated data & latent variable & original data\\
distance & Geodesic & Euclidean & Euclidean\\
'T-shirt'-'Scandal' & 1 & 0.98 & 0.48\\
'T-shirt'-'Scandal'-'Bag' & 1 & 0.93 & 0.28\\
\bottomrule
\end{tabular}
\end{table}

\subsubsection{The EMNIST-Letter Dataset}

\begin{figure}
  \centering
  \includegraphics[width=1.3in]{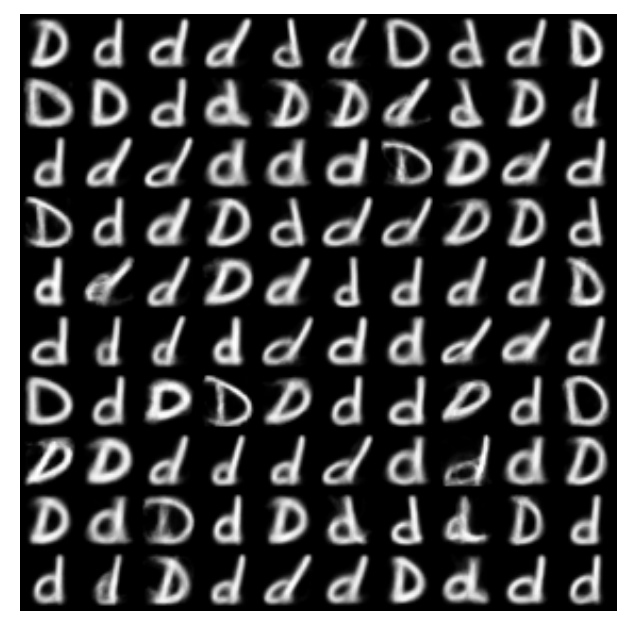}
  \caption{Generated 'D' and 'd' examples for EMNIST-letters by VAE used in the paper}\label{emnist_draw}
\end{figure}

The EMNIST-letter dataset \cite{emnist} is a set of handwritten alphabet characters
derived from the NIST Special Database and converted to $28\times 28$
gray-scale images. We select the characters 'D' and 'd' as 2 classes, and fit a VAE
with the same network architectures as the ones used for Fashion-MNIST.
Generated images are shown in Fig.~\ref{emnist_draw}.

\begin{figure*}
  \centering
  \subfigure[]{
    \includegraphics[width=2.0in]{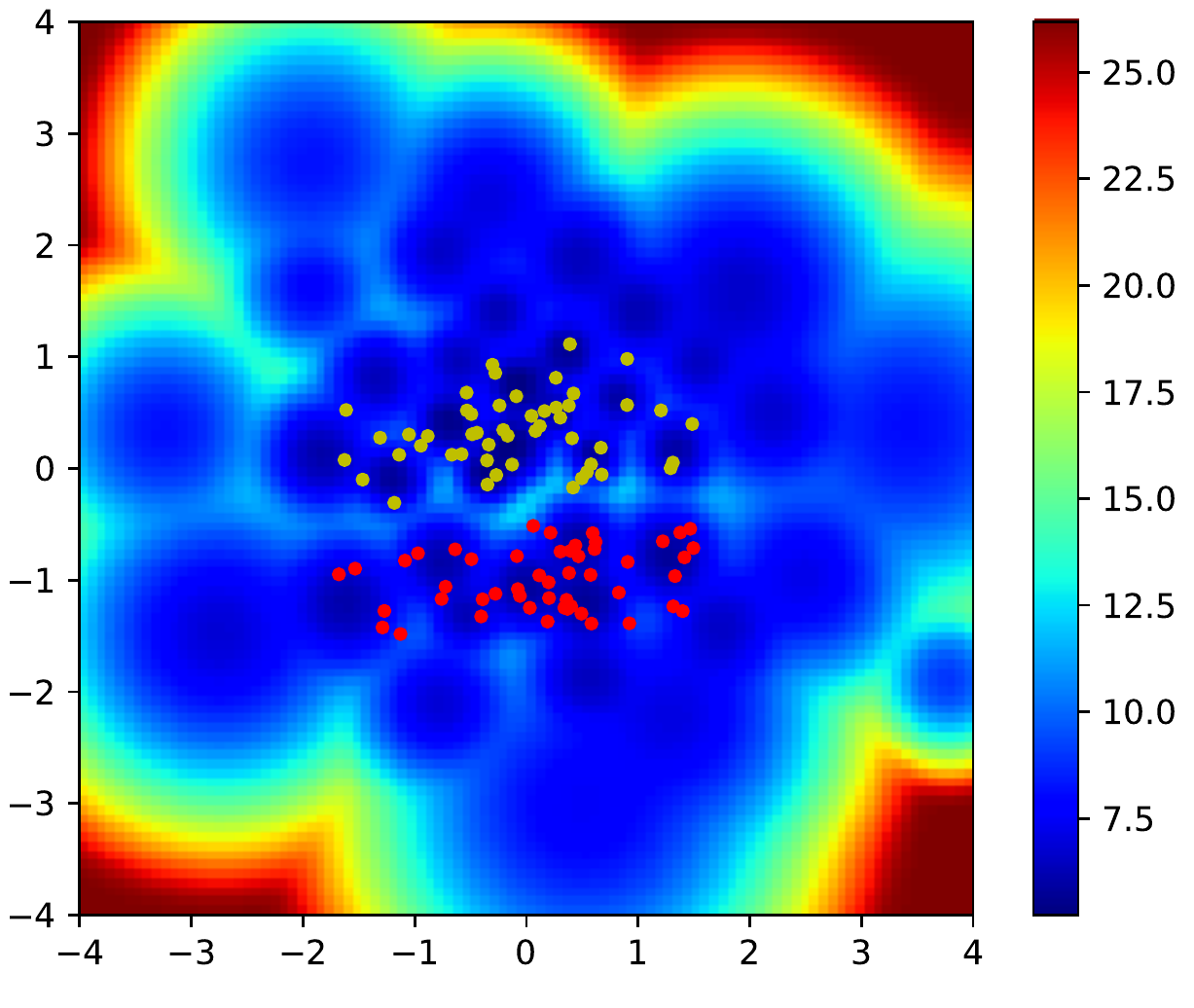}
  }
  \subfigure[]{
    \includegraphics[width=4.4in]{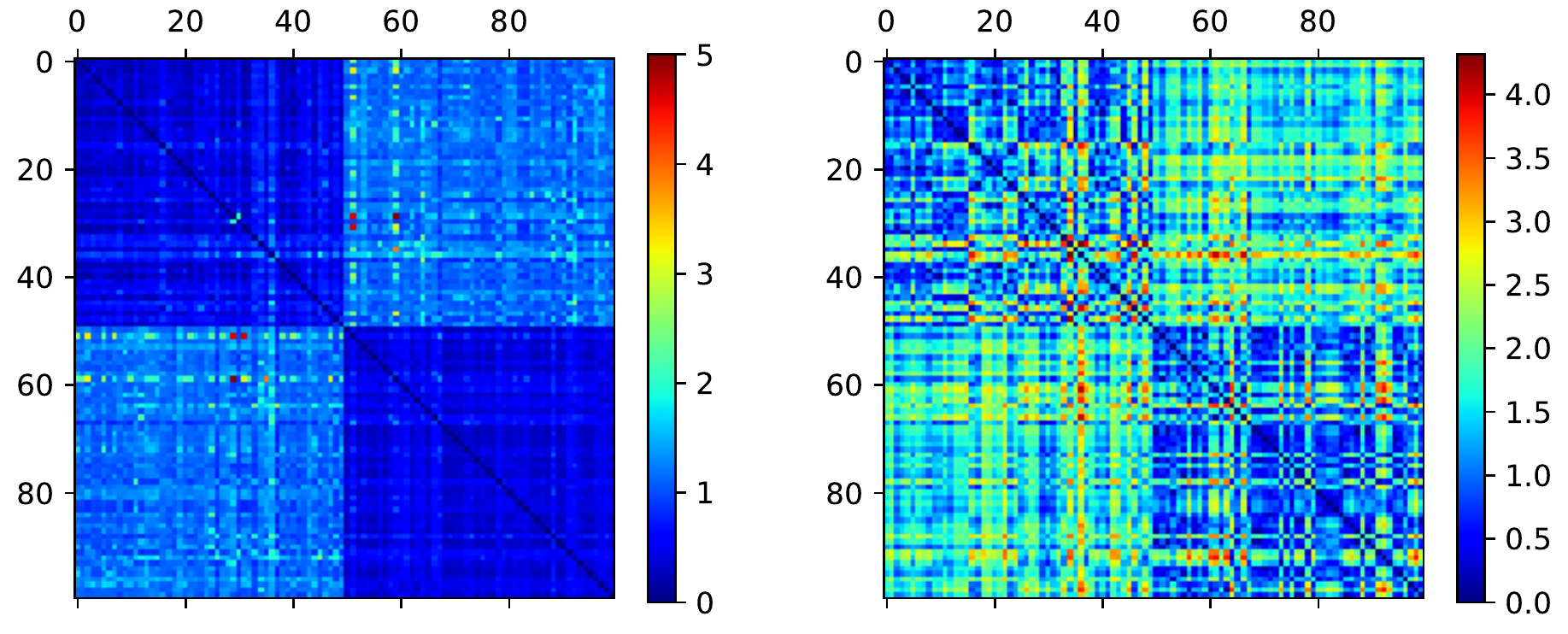}
  }
  \caption{(a): The logarithmic result of the area measure in latent space for 'D' and 'd' images. (b) Left: The optimized geodesic pair-wise distance. Right: The Euclidean pair-wise distance in latent space.}
  \label{emnist} 
\end{figure*}

We select 50 samples from 'D' and 'd' respectively and show pair-wise distances
in Fig.~\ref{emnist}. Again, $k$-medoids clustering show that the geodesic
distance reflects the intrinsic structure, which improves clustering over the baselines,
c.f.\ Table~\ref{emnist_tab}.


\begin{table}[!htbp]
\centering
\caption{EMNIST-letter dataset clustering 2-class accuracy}\label{emnist_tab}
\begin{tabular}{cccc}
\toprule
method & $k$-medoids & $k$-medoids & \textit{SC}\\
\midrule
data samples & reconstructed data & latent variable & original data\\
distance & Geodesic & Euclidean & Euclidean\\
'D'-'d' & 1 & 0.76 & 0.48\\
\bottomrule
\end{tabular}
\end{table}

\section{Conclusion}\label{sec:conclusion}
In this paper, we have proposed an efficient algorithm for computing shortest
paths (geodesics) along data manifolds spanned by deep generative models.
Unlike previous work, the proposed algorithm is easy to implement and fits well
with modern deep learning frameworks. We have also proposed a new network
architecture for representing variances in variational autoencoders.
With these two tools in hand, we have shown that simple distance-based
clustering works remarkably well in the latent space of a deep generative
model, even if the model is not trained for clustering tasks. Still, the dimension of the latent space, the form of the curve parametrization and modeling variance in generator worth developing further to obtain the more robust geodesics computation algorithm.


\section*{Acknowledgments}
TY was supported by the National Key R\&D Program of China (No. 2017YFB0702104). SH was supported by a research grant (15334) from VILLUM FONDEN. This project has
received funding from the European Research Council (ERC) under the European
Union's Horizon 2020 research and innovation programme (grant agreement n\textsuperscript{o} 757360).

\ifCLASSOPTIONcaptionsoff
  \newpage
\fi



\bibliographystyle{IEEEtran}
\bibliography{mybibfile}
\end{document}